\documentclass[10pt,twocolumn,letterpaper]{article}

\usepackage[pagenumbers]{cvpr} 

\usepackage[table]{xcolor}
\usepackage{colortbl}

\definecolor{cvprblue}{rgb}{0.21,0.49,0.74}
\usepackage[pagebackref,breaklinks,colorlinks,allcolors=cvprblue]{hyperref}
\usepackage{makecell}
\usepackage{booktabs,multirow}
\usepackage{times,graphicx,amsmath,amssymb,booktabs,mathtools,algorithm,algorithmic,xcolor}
\usepackage{color}
\definecolor{lm_purple_low}{RGB}{240,240,248}
\definecolor{lm_purple}{RGB}{227,227,240}
\definecolor{lm_purple_mid}{RGB}{233,222,254}
\definecolor{lm_red}{RGB}{230,36,43}
\definecolor{cblue}{rgb}{0.21,0.49,0.74}
\def\paperID{15387} 
\def\confName{CVPR}
\def\confYear{2026}

\title{R$^2$-Seg: Training-Free OOD Medical Tumor Segmentation \\ via Anatomical Reasoning and Statistical Rejection}

\author{%
  \textbf{Shuaike Shen}$^{1*}$, \textbf{Ke Liu}$^{3*}$, \textbf{Jiaqing Xie}$^4$, \textbf{Shangde Gao}$^3$,\\\textbf{Chunhua Shen}$^3$, \textbf{Ge Liu}$^5$, \textbf{Mireia Crispin-Ortuzar}$^2$, \textbf{Shangqi Gao}$^{2\dagger}$\\ 
$^1$Carnegie Mellon University, $^2$University of Cambridge,\\
$^3$Zhejiang University, $^4$ETH Zurich, $^5$University of Illinois Urbana-Champaign.\\
$^*$Equal contribution, $^\dagger$Corresponding authors\\
\texttt{shuaikes@andrew.cmu.edu, kliu@zju.edu.cn, sg2162@cam.ac.uk} \\
}

\begin{document}
\maketitle
\begin{abstract}

Foundation models for medical image segmentation struggle under out-of-distribution (OOD) shifts, often producing fragmented false positives on OOD tumors. We introduce \textbf{R$^2$-Seg}, a \textbf{training-free} framework for robust OOD tumor segmentation that operates via a two-stage \textbf{Reason-and-Reject} process. First, the \textbf{Reason} step employs an LLM-guided anatomical reasoning planner to localize organ anchors and generate multi-scale ROIs. Second, the \textbf{Reject} step applies two-sample statistical testing to candidates generated by a frozen foundation model (BiomedParse) within these ROIs. This statistical rejection filter retains only candidates significantly different from normal tissue, effectively suppressing false positives. Our framework requires no parameter updates, making it compatible with zero-update test-time augmentation and avoiding catastrophic forgetting. On multi-center and multi-modal tumor segmentation benchmarks, \textbf{R$^2$-Seg} substantially improves Dice, specificity, and sensitivity over strong baselines and the original foundation models. Code are available at \href{https://github.com/Eurekashen/R2Seg}{https://github.com/Eurekashen/R2Seg}.

\end{abstract}
\section{Introduction}
\label{sec:intro}
Cancer ranks as the second leading cause of death globally, and less invasive radiological imaging plays a central role in early screening, diagnosis, staging, treatment planning, and monitoring \cite{crosby2022science}. Consequently, accurate tumor segmentation across different organs or regions of the body is a critical first step in identifying radiological biomarkers for cancer detection and diagnosis. Recently, promptable image segmentation has progressed rapidly from SAM~\cite{kirillov2023segment}, MedSAM~\cite{ma2024segment} to text-driven \emph{BiomedParse}~\cite{zhao2024biomedparse}, which jointly performs segmentation, detection, and recognition without any expert intervention. However, tumors manifest as abnormal tissues with irregular shapes, varying sizes (ranging from millimeters to centimeters), and diverse intensity profiles, resulting in highly long-tailed distributions~\cite{zheng2024large}. In particular, tumors of the same type can exhibit substantial spatial heterogeneity across different scanners, imaging protocols, and patient populations~\cite{chen2024spatial}.
These factors could lead to pronounced out-of-distribution (OOD) shifts in medical images and increase high false positive rates in foundation segmentation models, potentially causing harmful overdiagnosis, heightened patient anxiety, and additional financial burden~\cite{yan2018deeplesion}.

A common approach to combat OOD shifts is fine-tuning~\cite{liang2025comprehensive}. However, collecting cancer imaging data remains challenging due to its scarcity, and tumor annotation is labor-intensive, requiring highly experienced experts. Fine-tuning foundation models on small tumor datasets may lead to catastrophic forgetting and diminish their generalization capacity. Recently, test-time adaptation (TTA) has emerged as a promising approach for adapting foundation models by only calibrating normalization layers. For instance, methods such as entropy minimization~\cite{wang2021tent} on batch normalization layers or test-time training with self-supervision~\cite{sun2020ttt} can alleviate distribution shift but often result in numerous small false positives in medical image segmentation. Furthermore, these approaches require access to the model architectures, which may not always be available.

These challenges lead to a critical question: \emph{Can a foundation model be adapted to OOD tumor segmentation without modifying its architecture or parameters?} To achieve that, we propose \textit{R$^2$-Seg}, a novel \textit{training-free} framework that affirmatively answers this question. R$^2$-Seg operates on a \textit{Reason-and-Reject} principle, enhancing OOD tumor segmentation by integrating external knowledge and statistical rejection without any parameter updates. The framework unfolds in two stages: First, the \textit{Reason} stage employs an LLM-guided anatomical reasoning planner. This planner translates a high-level goal (e.g., ``bladder tumor'') into an actionable, hierarchical plan: it identifies anatomical anchors (the organ) and generates constrained Regions of Interest (ROIs). The frozen foundation model is then prompted only within these localized ROIs.
Second, the \textit{Reject} stage addresses remaining false positives. We introduce a statistical two-sample testing filter that compares each potential candidate with normal tissue features. Only candidates that are statistically different from normal backgrounds are retained. This principled rejection process effectively suppresses fragmented artifacts.

The main contributions can be summarized as follows:
\begin{itemize}
    \item We introduce R$^2$-Seg, a novel training-free framework for OOD tumor segmentation that operates on a Reason-and-Reject principle.
    \item We design a Reason stage, where an LLM-guided planner performs anatomical reasoning to generate hierarchical ROIs.
    \item We propose a Reject stage, which employs principled two-sample statistical testing to filter false positives without any parameter updates.
    \item We demonstrate that our training-free framework avoids catastrophic forgetting and robustly outperforms both the original foundation model and strong baselines on multi-center and multi-modal OOD benchmarks.
\end{itemize}



\section{Related Works}
\label{sec:related_work}
\subsection{Medical image segmentation.}
Classical supervised approaches, including U-Net~\cite{ronneberger2015unet} and its variants~\cite{milletari2016vnet,cciccek2016unet3d,oktay2018attentionunet,isensee2021nnunet,liu2025towards}, have established the encoder-decoder architecture as a foundational paradigm in medical image segmentation. 
Based on this, nnU-Net~\cite{isensee2021nnunet} and TotalSegmentator~\cite{wasserthal2023totalsegmentator} have further advanced the field by developing automated configuration systems or leveraging anatomical priors for robust segmentation.
Meanwhile, the development of Transformer-CNN hybrid models, such as TransUNet~\cite{chen2024transunet} and Swin-UNet~\cite{cao2022swin}, further improved performance by enhancing global context modeling.
More recently, “foundation” and promptable segmentation models (e.g., MedSAM~\cite{ma2024segment}, SAM-Med2D~\cite{wu2023sammed2d}) have enabled flexible inference using text or point prompts. 
Specifically, BiomedParse~\cite{zhao2024biomedparse} unified segmentation, detection, and recognition tasks across nine imaging modalities through concise textual prompts.
While effective, BiomedParse tends to over-predict foreground regions in out-of-distribution (OOD) scenarios, leading to a significant increase in false positives.
To address this issue, we propose a test-time adaptation (TTA) framework that integrates reasoning-driven Region of Interest (ROI) planning and statistical hypothesis testing to enhance model robustness and reduce erroneous predictions.
\subsection{Test-time adaptation under distribution shift.}
Test-time adaptation (TTA)~\cite{sun2020test,liang2025comprehensive} improves model robustness to domain shifts using unlabeled test data. 
Pioneering self-supervised test-time training~\cite{sun2020ttt} and entropy minimization~\cite{wang2021tent} methods introduced lightweight parameter updates, establishing a foundation for efficient adaptation. 
EATA~\cite{niu2022eata} and CoTTA~\cite{wang2023cotta} further improved these strategies by refining online normalization and enhancing adaptation stability.
In medical segmentation, Karani \etal~\cite{karani2021testtime} proposed anatomy-guided normalization, and test-time augmentation~\cite{ayhan2018ttaug} leveraged multi-view fusion to reduce uncertainty.
Different from these parameter-updating approaches, our method keeps the model backbone frozen and integrates reasoning-guided region-of-interest (ROI) cropping, multi-view test-time fusion, and two-sample statistical filtering, explicitly designed to suppress false positives, particularly in empty-mask scenarios.

\subsection{LLM planning for anatomy-aware pipelines.}
Large language models (LLMs) are capable of decomposing complex goals and generating structured action plans through prompting and reasoning~\cite{li2023llavamed}. Concretely, chain-of-thought prompting~\cite{wei2022cot} and self-consistency~\cite{wang2022selfconsistency} enhance logical fidelity, while least-to-most prompting~\cite{zhouleast} enables the decomposition of intricate tasks. ReAct~\cite{yao2023react} combines reasoning with tool invocation, and approaches like Tree-of-Thoughts~\cite{yao2024tot} and Plan-and-Solve~\cite{wang2023planandsolve} formalize multi-step planning processes.
In our setting, an LLM planner translates a free-form medical concept (e.g., “bladder tumor”) into a structured \emph{AnatomyPlan} containing (i) anchor organs~$\mathcal{A}$, (ii) ROI instructions~$\mathcal{I}_{\text{ROI}}$, and (iii) a reasoning trace.  
This plan is later executed by a frozen segmentor (BiomedParse) to perform organ-based localization, ensuring prompts stay within the distribution of known anatomical entities while enabling compositional reasoning for unseen lesions.

\section{Methods}

\label{sec:method}
This work proposes a new test-time adaptation framework for OOD tumor segmentation. OOD tumor segmentation is very challenging due to the high heterogeneity of tumors across scanners, imaging protocols, and patient populations. To tackle the challenge, we combine anatomical reasoning, localized prompting, and statistical screening to mitigate high false positive issues in OOD tumor segmentation. Concretely, we illustrate the challenges of OOD test-time adaptation in Section \ref{sec:OOD_TTA}, and elaborate on the details of our test-time adaptation framework in Section \ref{sec:TTA_method}.

\begin{figure}[!htb]
    \centering
    \includegraphics[width=1\linewidth]{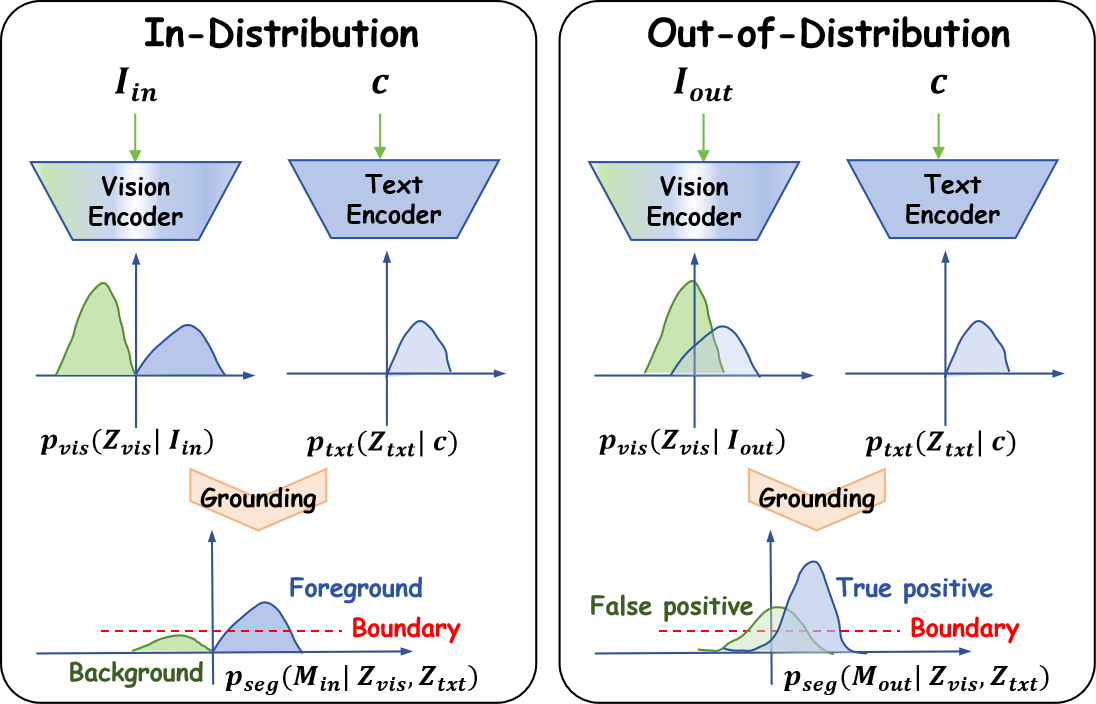}
    \caption{Illustration of visual embedding distributions. Left: In-Distribution, Right: Out-of-Distribution.}
    \label{fig:OOD_illustration}
\end{figure}

\subsection{OOD Test-time Adaptation}\label{sec:OOD_TTA}
Let $I_{in} \in \mathcal V_{in}$ denote an image from an in-distribution vision domain, $I_{out} \in \mathcal V_{out}$ denote an image from the out-of-distribution vision domain, and $c\in \mathcal C$ denote a textual prompt from a language domain. 
The vision encoder $p_{vis}(\cdot|I_{in})$ in conventional text-prompted foundation segmentation models is trained to map images into vision embeddings, $Z_{vis}$, and the language encoder $p_{txt}(\cdot|c)$ to map texts into language embeddings, $Z_{txt}$.
If vision embeddings are separable, these models can effectively distinguish between foreground and background by grounding text embeddings onto the foreground embeddings using a universal decision boundary, as shown in Fig.~\ref{fig:OOD_illustration} Left. 
However, medical imaging protocols vary across scanners, tumor sites, and modalities, which could lead to greatly different fields of view, artifacts, and noise. Thus, vision embeddings of OOD samples become hardly separable, and the decision boundary tends to be biased. Many background details would be recognized as tumors, leading to high false positive rates and harmful over-diagnosis, as shown in Fig.~\ref{fig:OOD_illustration} Right.

\begin{figure*}
    \centering
    \includegraphics[width=0.85\linewidth]{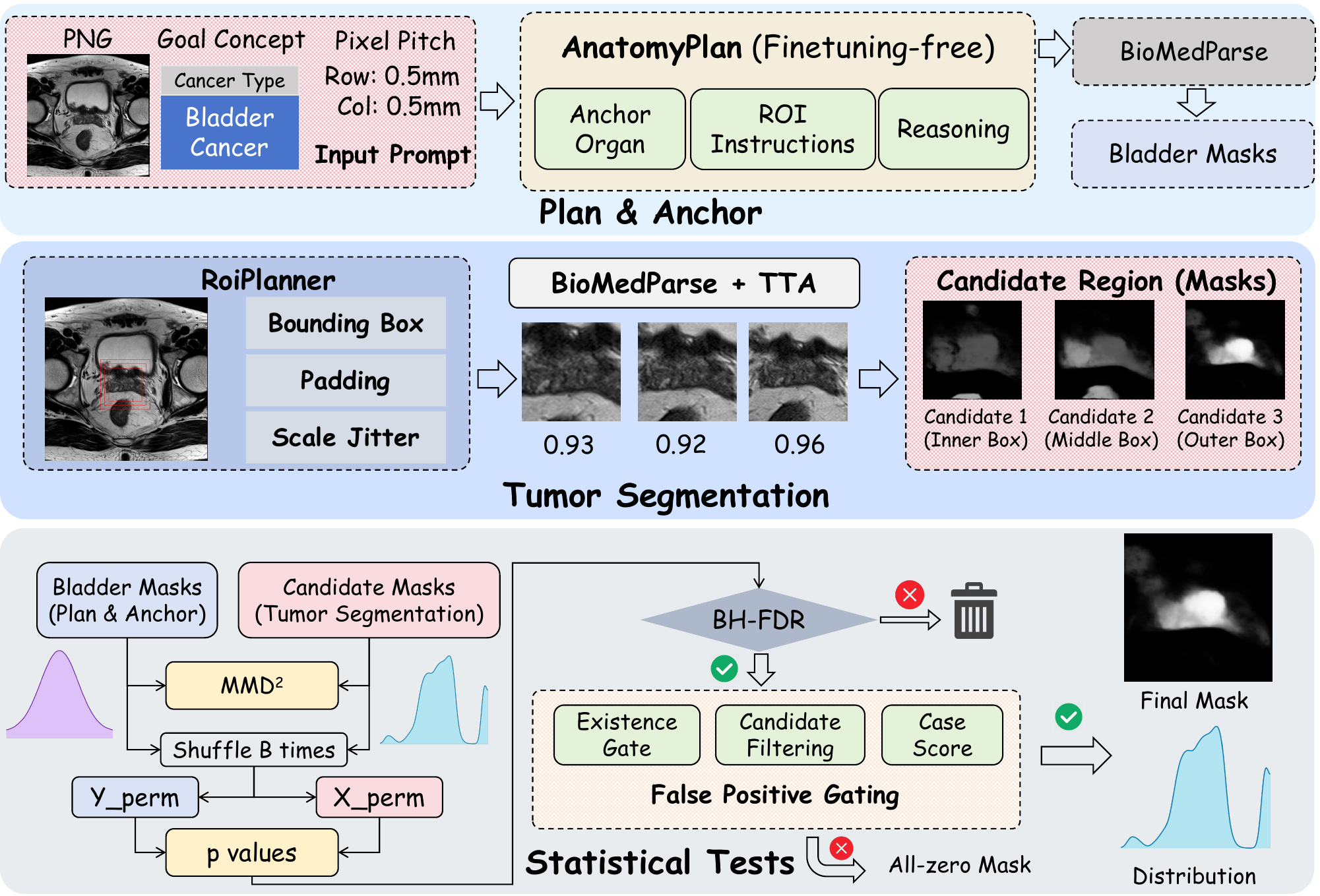}
    \caption{Overview of R$^2$-Seg pipeline. Top row: LLM-based segmentation planning and ROI construction; middle row: BioMedParse-based tumor segmentation and candidate extraction; bottom row: Statistical two-sample test and false discovery rate control.}
    \label{fig:tta_pipeline}
\end{figure*}

To mitigate the over-diagnosis problem, we propose a new training-free test-time adaptation framework for foundation segmentation models by anatomy-aware reasoning and statistical rejection, as shown in Fig.~\ref{fig:tta_pipeline}. 
This framework consists of two main components: (1) \textit{enhancing the separability} of vision embeddings by reasoning to plan regions of interest (ROIs), and (2) \textit{calibrating the decision boundary} through statistical screening of normal and abnormal regions, which together address the core challenges of OOD test-time adaptation. 
For the former, we use LLM to plan possible anchors, margins, planes, and scales of normal anatomical organs surrounding a given tumor site. 
Then, we adopt foundation segmentation models to segment normal organs. Based on the masks of normal organs, we generate ROIs covering the tumor site from different views. For the latter, we segment tumors within each ROI using text-prompted foundation segmentation models for ensembling. 
However, high false positive rates are observed in tumor segmentation due to the imaging artifacts and noise. 
Therefore, we further introduce statistical rejection for normal and abnormal regions, which mitigates overdiagnosis by rejecting tumor candidates with low confidence. 

\subsection{\texorpdfstring{R$^2$-Seg pipeline overview}{R2-Seg pipeline overview}}
\label{sec:TTA_method}

\subsubsection{LLM-based Planning and ROI Construction}
 
We use an LLM planner to reason about the topology of the cross-organ and propose region-of-interest (ROI) instructions that improve the separability of target anatomy (Fig.~\ref{fig:tta_pipeline}, top).
Formally, the planner $\Phi$ maps the textual cancer type $c$ to
\[
\Phi(c) \rightarrow (\mathcal{A}, \mathcal{I}_{\mathrm{ROI}}, r),
\]
where $\mathcal{A}$ denotes a set of \emph{anchor organs}, and $\mathcal{I}_{\mathrm{ROI}}$ encodes geometric rules, including padding $\delta$, scale jitter set $\Gamma$, and square enforcement. 
Given a text-conditioned segmentor $f_{\theta}(\cdot)$ and a threshold $\tau_a$ (default $0.5$), for each $a\in\mathcal{A}$ we obtain a probability map $P_a = f_{\theta}(I; c_a, \tau_a)$ and a binary mask $M_a=\mathbb{1}\{P_a\ge \tau_a\}$, where $c_a$ is the anchor-specific prompt. 
Let $M_{\cup}=\bigcup_{a\in\mathcal{A}} M_a$ and $B_0=\mathsf{BBox}(M_{\cup})$ be the union mask and its axis-aligned bounding box, respectively. 
We then define padded, square ROIs by
\begin{equation}
B_{\gamma}=\mathsf{Square}\!\big(\mathsf{Dilate}(B_0,\ \lceil\delta/s\rceil\cdot\gamma)\big),\quad \gamma\in\Gamma,
\end{equation}
where $s$ is the in-plane pixel spacing (in pixels per mm), $\mathsf{Dilate}$ enlarges $B_0$ by a given margin, and $\mathsf{Square}$ enforces a square crop. 
Each $B_{\gamma}$ induces a cropped input $I|_{B_{\gamma}}$ for subsequent inference. More details can be found in App.~\ref{app:method_details}.

\subsubsection{Tumor Segmentation and Candidate Extraction}
For each ROI, the frozen segmentor performs multi-view test-time augmentation and \emph{max-fuses} the predictions back to the native spatial resolution:
\begin{equation}
\bar{P} = \max_{g \in \mathcal{G}}
\big[ \mathsf{Inv}(g) \circ f_{\theta}\big(g(I|_{B_{\gamma}}); c_{\mathrm{tumor}}, \tau_{\mathrm{tumor}} \big) \big],
\end{equation}
where $c_{\mathrm{tumor}}$ denotes the tumor-specific prompt, and $\mathcal{G} = \lbrace g_{\mathrm{id}}, g_{\mathrm{lr}}, g_{\mathrm{tb}}\rbrace$ represents the set of geometric transformations including identity, left–right flip, and top–bottom flip. The inverse mapping $\mathsf{Inv}(g)$ restores predictions to the original coordinate frame.
A binary segmentation mask $\mathcal{M}_{\mathrm{tumor}} = \mathbb{1}\lbrace \bar{P} \ge \tau_{\mathrm{tumor}}\rbrace$ is then obtained via thresholding, from which a collection of spatially disjoint connected regions $\lbrace C_k \rbrace_{k \in \mathcal{K}}$ is extracted by applying a connected-component decomposition operator $\mathsf{Conn}(\cdot)$:
\begin{equation}
\lbrace C_k \rbrace_{k \in \mathcal{K}} = \mathsf{Conn}(\mathcal{M}_{\mathrm{tumor}}).
\end{equation}

Additional implementation details are provided in Algorithm.~\ref{alg:seg}.

\subsubsection{Statistical Two-sample Test and FDR Control}
\label{sec:stats}
Learning to reject false positives and reshape decision boundary is critical for tumor segmentation. Benefit from the anatomical reasoning and normal organ segmentation, we screen radiological features of both normal organs and abnormal regions, and statistically test their differences, as shown in Fig~\ref{fig:tta_pipeline} Bottom. 
Before the two-sample test, we first generate candidates by thresholding at $\tau_{\mathrm{bin}}=0.4$ for connected-component extraction and removing very small components (area $\ge 50$ px).
Concretely, for each tumor component $C_k$, its pixel-level feature is compared against that of the normal organ mask $M_{\cup}$ via a nonparametric two-sample test.  
Let feature extractor $\phi(\cdot)$ (e.g., normalized intensity) produce sets  
$X=\{\phi(I|_{C_k})\}_{i=1}^m$ and $Y=\{\phi(I|_{M_{\cup}})\}_{j=1}^n$, where $m=|C_k|$ and $n=|M_{\cup}|$.  
Then, we test the following null assumption:
\[
H_0:P_X=P_Y,\qquad H_1:P_X\neq P_Y.
\]
Under the null assumption, using the unbiased squared maximum mean discrepancy (MMD$^2$)~\cite{gretton2012kernel} with Gaussian kernel $k_\sigma(u,v)=\exp(-\|u-v\|_2^2/2\sigma^2)$, i.e.,
\begin{align}
\widehat{\mathrm{MMD}}^2
   &=\tfrac{1}{m(m-1)}\!\sum_{i\neq i'}k_\sigma(x_i,x_{i'})
     +\tfrac{1}{n(n-1)}\!\sum_{j\neq j'}k_\sigma(y_j,y_{j'})\nonumber\\
   &\quad-\tfrac{2}{mn}\!\sum_{i,j}k_\sigma(x_i,y_j),
\end{align}
 and randomly permuting the combined samples $\lbrace X, Y\rbrace$ with $B$ times, i.e., $\lbrace X_{\mathrm{perm}, b},  Y_{\mathrm{perm}, b}\rbrace_{b=1}^B$, we estimate the permutation $p$-value of the $k$-th candidate by
\[
p_k=\tfrac{\left|\lbrace b|\widehat{\mathrm{MMD}}^2_{\text{perm},b}\!\ge\!\widehat{\mathrm{MMD}}^2_{\text{obs}}\rbrace\right|\!+\!1}{B+1},
\]
where, $\widehat{\mathrm{MMD}}^2_{\text{obs}}$ denotes the observed statistic between the actual $X$ and $Y$, and $\widehat{\mathrm{MMD}}^2_{\text{perm},b}$ denotes the permuted statistic between $X_{\mathrm{perm},b}$ and $Y_{\mathrm{perm},b}$.
Across $|\mathcal K|$ candidates, we use Benjamini–Hochberg correction~\cite{benjamini1995fdr} to control False Discovery Rate (FDR)  at level $\alpha$, including (1) 
sorting the $p$-value,  $p_{(1)}\!\le\!\cdots\!\le\!p_{(|\mathcal K|)}$, (2) searching a decision boundary, $i^\ast=\max\{i:p_{(i)}\!\le\!\alpha i/|\mathcal K|\}$, and (3) keeping $\{C_{(1)},\dots,C_{(i^\ast)}\}$ as significant anomalies.  
Without loss of generality, other metrics (e.g., energy distance~\cite{szekely2013energy}) may replace MMD for statistical tests.
We further subsample each class to at most $4\mathrm{k}$ samples and set $\sigma$ via the median heuristic; $p$-values use the smoothed estimate $(\mathrm{count}+1)/(B+1)$.

\subsubsection{False-positive Gating for Empty-mask Cases}
\label{sec:fp_mitigation}
Text-prompted segmentors rarely output empty masks, producing high false-positive rates when images do not contain tumors. To reject predictions with low confidence, we introduce a three-level gating policy:

\noindent\textbf{(L1) Existence gate.}
Compute the global maximum $p_{\max}=\max_x P(x)$, positive ratio $\rho$, and Kolmogorov–Smirnov $p_{\text{KS}}$ by comparing foreground and background probabilities. 
Here $P(x)$ denotes the \emph{pixel-wise max-fused probability map} aggregated across TTA views, prompts, and both ROI and full-image inputs; $\rho$ is computed \emph{within the union ROI region} to avoid diluting small lesions with large backgrounds; the KS test uses \emph{foreground = organ control region} and \emph{background = its complement}. 
If $p_{\max}<\tau_{\max}$ or $\rho<\tau_{\rho}$ or $p_{\text{KS}}>\tau_{\text{KS}}$, the image is declared negative.

\noindent\textbf{(L2) Candidate-level gate.}
Filter surviving candidates by constraining area, mean probability $\overline{P}_k$, and overlap ratio with the organ mask through
$|C_k|\!\ge\!A_{\min}$, $\overline{P}_k\!\ge\!\tau_{\text{mean}}$, and $|C_k\!\cap\!M_{\cup}|/|C_k|\!\ge\!\tau_{\cap}$ respectively.

\noindent\textbf{(L3) Case-level score.}
For the remaining candidates, define $S_k=\overline{P}_k\sqrt{|C_k|}$ and $S^\ast=\max_k S_k$.  
If $S^\ast<\tau_{\text{case}}$, output an all-zero mask.  
This acts as a conservative post-hoc calibration that bounds the false-positive rate when empty masks dominate the test set.

\subsubsection{Complexity and Guarantees}
Permutation-based $p$-values ensure that, under $H_0$, the expected false discovery rate is less than $\alpha$ by the BH theorem~\cite{benjamini1995fdr}.  
The computational complexity is $O(|\mathcal K|\cdot B\cdot (m+n)^2)$, and fast MMD approximations can reduce cost without altering guarantees.

\section{Experiments}

\subsection{Datasets and splits}
We evaluate our approach on ten organ-specific tumor segmentation datasets spanning both computed tomography (CT) and magnetic resonance (MR) imaging modalities, with statistics detailed in Table~\ref{tab:dataset_stats}.
The datasets encompass a diverse range of anatomical regions and imaging planes, including axial, sagittal, and mixed orientations, ensuring a comprehensive representation of tumor types and scanning conditions.
Each dataset is partitioned at the patient level into training (80\%) and test (20\%) subsets, with no patient overlap between splits. 
From the training scans, only slices containing tumors are extracted to train the baseline segmentation models.
Notably, our method operates under a \textit{training-free test-time adaptation} paradigm, and all model adaptation procedures are performed exclusively on the held-out test slices.
For evaluation, test slices are sampled with a balanced tumor-to-non-tumor ratio (1:1) to ensure fair comparison across tumor types and modalities.
\begin{table}[htbp]
    \centering
    \resizebox{1\linewidth}{!}{
    \begin{tabular}{c|c|c|c|c|c}
    \hline
    Dataset&  Modality& In-plane& Scans& Train slices& Test slices\\
    \hline
    \multicolumn{6}{c}{\cellcolor{lm_purple_low} Out-of-distribution tumor types}\\
    \hline
    Bladder tumor~\cite{bladdertumor}& MR&  Ax& 221&  1,194&  496\\ 
    Uterus tumor~\cite{uterustumor}& MR& Sag& 300& 2,094& 660\\
    Prostate tumor~\cite{prostatetumor}& MR& Ax& 316& 933& 656\\
    Breast tumor~\cite{breasttumor}& MR& Ax, Sag& 1,506& 31,514& 1,690\\
    Cervix tumor~\cite{cervixtumor}& MR& Sag& 67& 428& 234\\
    \hline
    \multicolumn{6}{c}{\cellcolor{lm_purple_low} In-distribution tumor types}\\
    \hline
    Liver tumor~\cite{MSD}& CT& Ax& 201& 5,043& 1,158\\
    Lung tumor~\cite{lungtumor,MSD}& CT& Ax& 484& 9,901& 1,740\\
    Pancreas tumor~\cite{MSD}& CT& Ax& 500& 2,079& 856\\
    Colon tumor~\cite{MSD}& CT& Ax& 126& 1,049& 472\\
    Kidney tumor~\cite{kidneytumor}& CT& Ax& 486& 3,855& 1,054\\
    \hline
    \multicolumn{6}{c}{\cellcolor{lm_purple_low} In-distribution data for evaluating forgetting}\\
    \hline
    AMOS22 CT~\cite{AMOS22}& CT& Ax& -& -&  29,664\\
    AMOS22 MR~\cite{AMOS22}& MR& Ax& -& -&  8,150\\
    M\&Ms~\cite{MMSheart}& MR& Cardiac& -& -&  1,792\\
    \hline
    \end{tabular}
    }
    \caption{A summary of datasets for tumor segmentation. Ax and Sag refer to Axial and Sagittal planes respectively.}
    \label{tab:dataset_stats}
\end{table}

\subsection{Baselines and variants}
We compare our R$^2$-Seg with commonly used techniques to adapt models to unseen domains.
(1) \textit{Lower bound}: the official \textbf{BiomedParse} model applied directly in a zero-shot manner.
(2) \textit{Upper bound}: directly fine-tuned BiomedParse using training slices in Table~\ref{tab:dataset_stats}, \textbf{BiomedParse-FT}. (3) \textit{Commonly} used approach: 
fine-tuned BioMedParse on tumor slices while applying low-rank adaptation (LoRA)~\citep{hu2022lora} to the pixel decoder and keeping the backbone frozen, \textbf{BioMedParse-LoRA}.

All these methods, including our R$^2$-Seg are evaluated on the same splits.
\subsection{Metrics and implementation}

Both \textit{pixel-level} and \textit{slice-level} metrics are reported to provide a comprehensive assessment of segmentation performance.
At the pixel level, \textbf{accuracy}, \textbf{F1-score} (equivalent to hard Dice), \textbf{Dice}, \textbf{soft Dice}, and \textbf{class-average accuracy (CA)} are employed.
Accuracy is defined as the proportion of correctly classified pixels.
The F1-score and Dice are computed from binary masks and are assigned a value of 1 when both masks are empty.
Soft Dice is calculated from probabilistic predictions with $\varepsilon$-smoothing and is similarly set to 1 in all-empty cases.
CA is obtained as the mean of positive and negative class accuracies to alleviate the impact of class imbalance.

To assess clinical relevance, \textbf{sensitivity} and \textbf{specificity} are computed at the \textit{slice level}, where pixel-level metrics may be unreliable due to the sparse tumor area.
Let $N_{\mathrm{pos}}$ and $N_{\mathrm{neg}}$ denote the numbers of positive and negative slices, respectively.
A slice is considered positive if any pixel within the predicted mask is classified as tumor.
Based on this criterion, the slice-level sensitivity and specificity are defined as:
\begin{equation}
\mathrm{Sensitivity} = \frac{N_{\mathrm{TP}}}{N_{\mathrm{TP}} + N_{\mathrm{FN}}},
\mathrm{Specificity} = \frac{N_{\mathrm{TN}}}{N_{\mathrm{TN}} + N_{\mathrm{FP}}},
\end{equation}
where $N_{\mathrm{TP}}$, $N_{\mathrm{FN}}$, $N_{\mathrm{TN}}$, and $N_{\mathrm{FP}}$ are the numbers of true-positive, false-negative, true-negative, and false-positive slices, respectively. 
Sensitivity is defined as 0 when there are no positive slices, and specificity is 1 when all slices are negative. 

\paragraph{Implementation details.}
R$^2$-Seg is a \textit{training-free} framework that adapts to domain shifts solely through test-time parameter adjustment.
All hyperparameters are organized into three functional groups: \textit{scoring}, \textit{statistical}, and \textit{geometric}.
The scoring group adopts max-based test-time augmentation (TTA) aggregation to retain high-confidence activations under mild scale perturbations $([0.8,1.0,1.2])$, with an adaptive binarization threshold $\tau_{\mathrm{bin}}\!\in\![0.30,0.55]$ ($\tau_{\mathrm{bin}}{=}0.4$ by default) balancing Dice and false positives.
The statistical group regulates \textit{Type-I error} via Gaussian kernel testing with Benjamini–Hochberg correction ($\alpha\!\in\!{0.05,0.01}$), enabling selective suppression of uncertain regions.
The geometric group enforces anatomical plausibility through ROI-normalized gating thresholds ($\tau_{\max}{=}0.45$, $\tau_{\mathrm{area}}{=}80$ px, etc.).
Together, these grouped parameters enable reasoning-guided adaptation that stabilizes predictions across unseen domains without retraining. More implementation details are referred to supplement \ref{supply:implementation}.  

\subsection{Results}
\label{sec:results}


\begin{figure*}
    \centering
    \includegraphics[width=0.93639\linewidth]{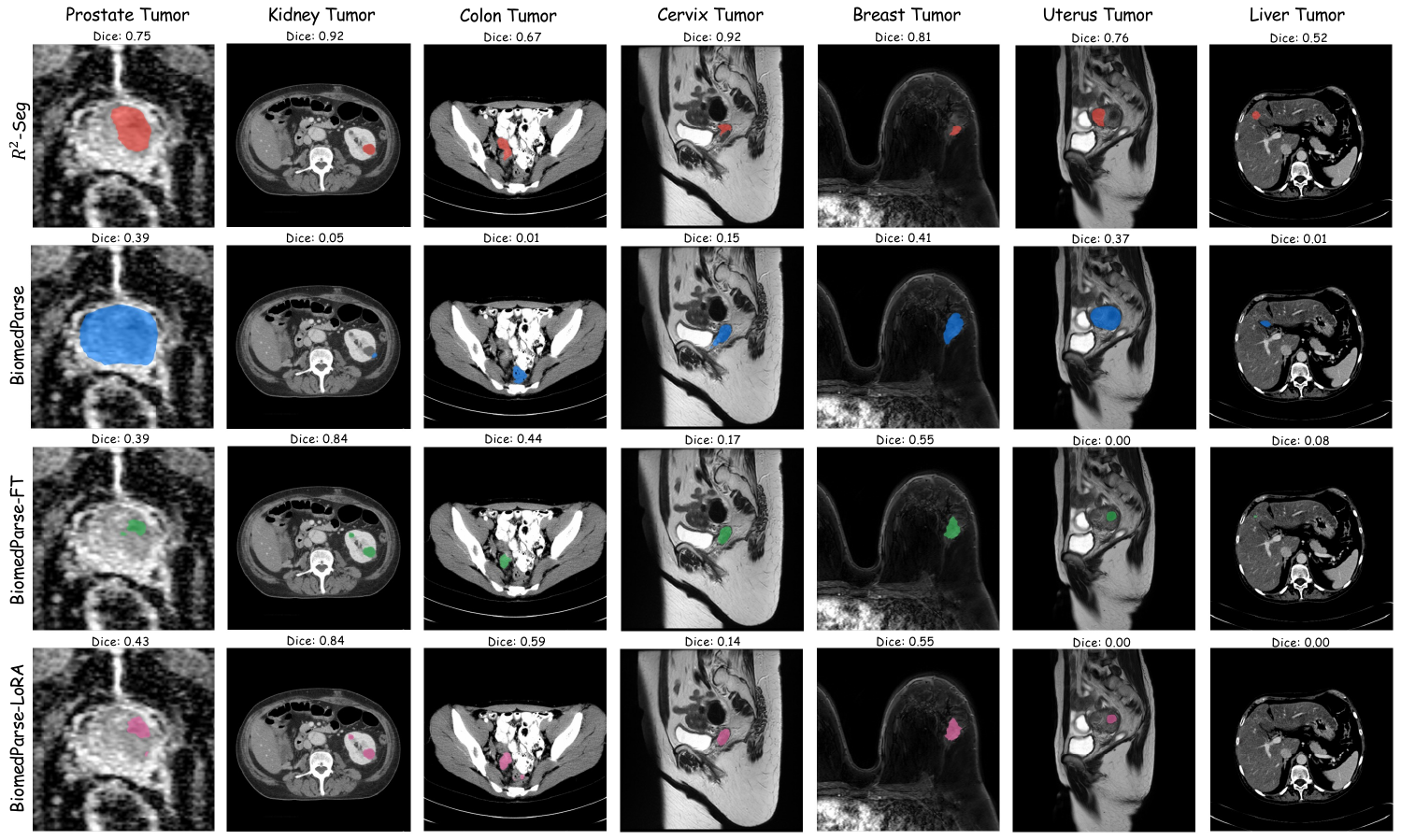}
    \vspace{-0.82494em}
    \caption{Visualization of segmentation results for both in-distribution and out-of-distribution tumor types.}
    \label{fig:exp_vis}
\end{figure*}

\begin{figure}
    \centering
    \includegraphics[width=1\linewidth]{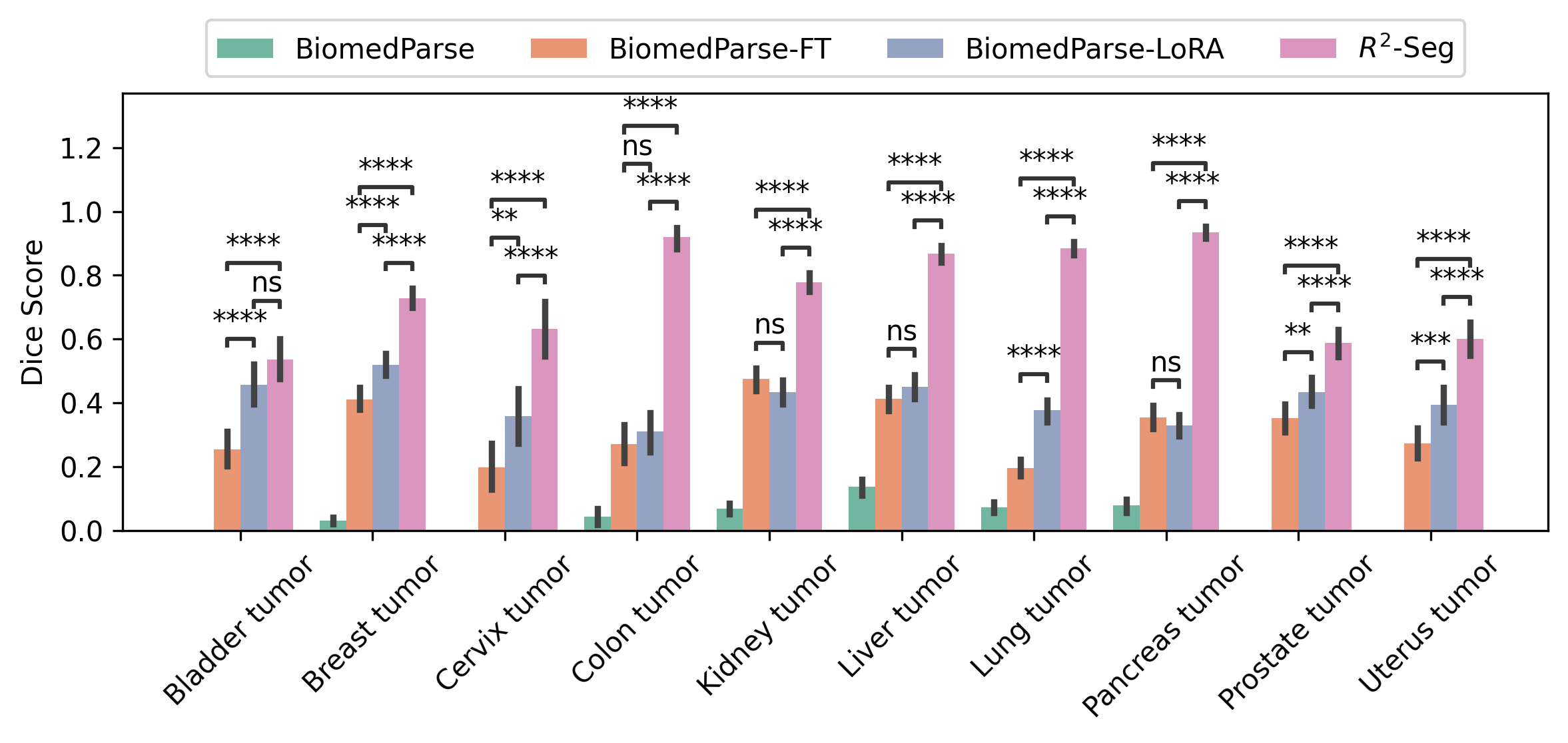}
    \vspace{-2.4em}
    \caption{Evaluation of over-segmentation on slices without tumors. Here, $p$-value is annotated by an asterisk, i.e., ns: $0.05 < p \le 1 $, *: $0.01 < p \le 0.05$, **: $0.001 < p \le 0.01$, ***: $0.0001 < p \le 0.001$, and ****: $ p \le 0.0001$.}
    \label{fig:background_seg}
\end{figure}

\begin{figure*}
    \centering
    \includegraphics[width=1\linewidth]{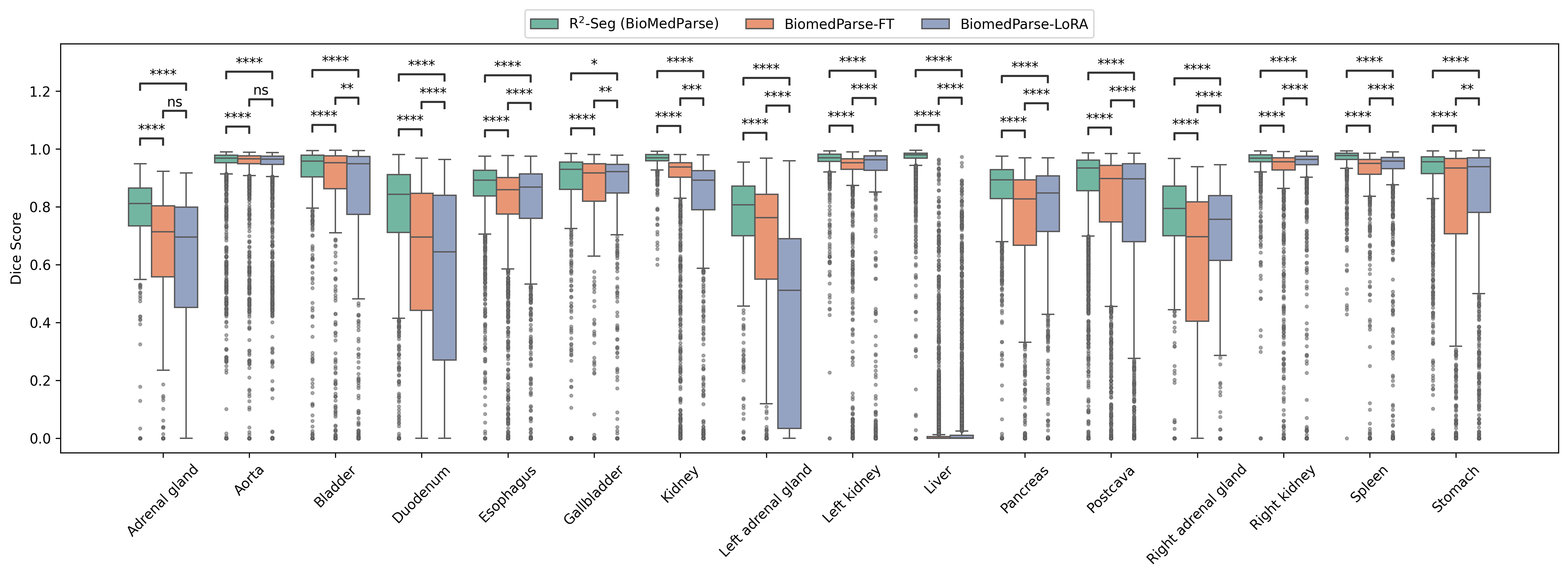}
    \vspace{-0.8em}
    \caption{Evaluation of knowledge forgetting on in-distribution CT slices. Statistical tests show that the segmentation performance of fine-tuned models drops significantly across all organs.}
    \label{fig:CT_abdomen_seg}
\end{figure*}


\subsubsection{Quantitative Comparison Across All Slices}
Table~\ref{tab:tumor_metrics_subset} summarizes the quantitative results on all evaluation slices, including both tumor and tumor-free cases.  
Across nearly all tumor types, R$^2$-Seg consistently outperforms both BiomedParse and BiomedParse-LoRA in overall accuracy and balance between sensitivity and specificity.  
While BiomedParse-LoRA provides modest improvements in Dice through limited low-rank fine-tuning, its adaptation remains largely local and often fails to generalize beyond the fine-tuned distribution.  
In contrast, R$^2$-Seg achieves substantially higher specificity and class-average accuracy, reflecting more stable and calibrated decision boundaries under domain shift.  
Notably, for liver and pancreas tumors, two of the most challenging cross-domain cases, R$^2$-Seg yields 10--30\% relative gains in both Dice and CA, while maintaining nearly perfect pixel-level accuracy. Moreover, when combing Table~\ref{tab:tumor_metrics_subset} with Fig.~\ref{fig:exp_vis}, it is noteworthy that BiomedParse tends to segment the entire organs (tumor site) containing tumor rather than the tumor itself in OOD cases such as prostate, cervix, uterus, and bladder. As a result, it achieves 100\% sensitivity but zero specificity, revealing the limited separability of its visual embeddings in OOD scenarios. In contrast, R$^2$-Seg leverages anatomical reasoning to plan ROIs more precisely, thereby enhancing the separability of its vision embeddings.
These results indicate that our \textit{test-time adaptation} mechanism not only suppresses noise from irrelevant regions but also preserves lesion integrity, demonstrating strong generalization to unseen tumor morphologies and modalities.  

\begin{table}[!ht]
\centering
\small
\setlength{\tabcolsep}{2pt}
\begin{tabular}{l l c c c c c}
\toprule
Tumor & Method & Dice & Sens. & Spec. & Acc. & CA \\
\midrule
\multirow{3}{*}{Bladder} 
& BiomedParse & 0.069 & 1.000 & 0.000 & 0.976 & 0.546 \\
& BiomedParse-LoRA & 0.578 & 0.960 & 0.456 & 0.996 & 0.677 \\
& R$^2$-Seg (Ours) \cellcolor{lm_purple_low} & \textbf{0.297} \cellcolor{lm_purple_low} & 0.335 \cellcolor{lm_purple_low} & \textbf{0.536} \cellcolor{lm_purple_low} & 0.992 \cellcolor{lm_purple_low} & \textbf{0.762} \cellcolor{lm_purple_low} \\
\midrule
\multirow{3}{*}{Uterus} 
& BiomedParse & 0.190 & 1.000 & 0.000 & 0.977 & 0.632 \\
& BiomedParse-LoRA & 0.514 & 0.897 & 0.394 & 0.995 & 0.670 \\
& R$^2$-Seg (Ours) \cellcolor{lm_purple_low} & \textbf{0.401} \cellcolor{lm_purple_low} & 0.424 \cellcolor{lm_purple_low} & \textbf{0.600} \cellcolor{lm_purple_low} & 0.989 \cellcolor{lm_purple_low} & \textbf{0.806} \cellcolor{lm_purple_low} \\
\midrule
\multirow{3}{*}{Prostate} 
& BiomedParse & 0.047 & 1.000 & 0.000 & 0.910 & 0.552 \\
& BiomedParse-LoRA & 0.428 & 0.852 & 0.434 & 0.992 & 0.555 \\
& R$^2$-Seg (Ours) \cellcolor{lm_purple_low} & \textbf{0.465} \cellcolor{lm_purple_low} & 0.645 \cellcolor{lm_purple_low} & \textbf{0.587} \cellcolor{lm_purple_low} & 0.971 \cellcolor{lm_purple_low} & \textbf{0.890} \cellcolor{lm_purple_low} \\
\midrule
\multirow{3}{*}{Breast} 
& BiomedParse & 0.144 & 0.998 & 0.030 & 0.952 & 0.582 \\
& BiomedParse-LoRA & 0.611 & 0.954 & 0.520 & 0.996 & 0.685 \\
& R$^2$-Seg (Ours) \cellcolor{lm_purple_low} & \textbf{0.395} \cellcolor{lm_purple_low} & 0.412 \cellcolor{lm_purple_low} & \textbf{0.728} \cellcolor{lm_purple_low} & 0.984 \cellcolor{lm_purple_low} & \textbf{0.762} \cellcolor{lm_purple_low} \\
\midrule
\multirow{3}{*}{Cervix} 
& BiomedParse & 0.154 & 1.000 & 0.000 & 0.985 & 0.598 \\
& BiomedParse-LoRA & 0.485 & 0.949 & 0.359 & 0.996 & 0.686 \\
& R$^2$-Seg (Ours) \cellcolor{lm_purple_low} & \textbf{0.355} \cellcolor{lm_purple_low} & 0.299 \cellcolor{lm_purple_low} & \textbf{0.632} \cellcolor{lm_purple_low} & 0.993 \cellcolor{lm_purple_low} & \textbf{0.777} \cellcolor{lm_purple_low} \\
\bottomrule
\end{tabular}
\vspace{-0.41257em}
\caption{Representative results across five OOD tumor types. Mean values of Dice, sensitivity (Sens.), specificity (Spec.), accuracy (Acc.), and class-average accuracy (CA) are reported.}
\label{tab:tumor_metrics_subset}
\end{table}

\subsubsection{Comparison on Background Slices}
When evaluated on tumor-free slices, R$^2$-Seg shows a significant improvement in controlling false activations.  
Since BiomedParse is trained primarily on in-distribution anatomical structures and rarely outputs empty masks, it produces dense spurious activations on out-of-distribution or background slices, resulting in near-zero Dice scores for several tumor categories, such as bladder, cervix, prostate, and uterus.  
Although BiomedParse-LoRA partially alleviates this issue by adjusting feature activations through fine-tuning, it often overcompensates, activating irrelevant regions and disrupting structural consistency.
By contrast, the hierarchical \textit{false-positive gating} and two-sample \textit{statistical filtering} modules in R$^2$-Seg adaptively regulate prediction confidence and spatial sparsity during inference, effectively removing background responses without retraining.  
As a result, R$^2$-Seg produces near-empty masks on negative slices while maintaining reliable recall on true lesions, demonstrating its robustness and calibration advantage under severe distribution shifts.

\subsubsection{Summary of Findings}
Across both evaluation regimes, R$^2$-Seg  delivers robust test-time performance improvements by coupling LLM-guided anatomical reasoning with lightweight statistical adaptation.
The results validate the effectiveness of \emph{planning before segmentation} and highlight that substantial gains in OOD robustness can be achieved without any fine-tuning of the segmentation model. Furthermore, the evaluation results in Figure~\ref{fig:CT_abdomen_seg} and supplementary Figure~\ref{fig:MR_abdomen_seg} show that the finetuned models present catastrophic forgetting on both CT and MR abdominal organ segmentation, especially on liver segmentation, due to the sparsity nature of tumors compared with normal organs. In contrast, our R$^2$-Seg does not require updating the model weights of BiomedParse, thereby avoiding the issue of knowledge forgetting.

\begin{table}[!ht]
\centering
\small
\setlength{\tabcolsep}{4pt}
\resizebox{1\linewidth}{!}{
\begin{tabular}{l c c c c c c c}
\toprule
Tumor & \makecell{Stat.\\Test} & \makecell{FP\\Gating} & Dice & Sens. & Spec. & Acc. & CA \\
\midrule
\multirow{3}{*}{Bladder}
& \checkmark & & 0.099{\tiny$\pm$0.27} & \textbf{0.923} & 0.089 & 0.989{\tiny$\pm$0.02} & 0.774{\tiny$\pm$0.24} \\
& & \checkmark & 0.257{\tiny$\pm$0.42} & 0.540 & 0.419 & 0.990{\tiny$\pm$0.02} & 0.770{\tiny$\pm$0.24} \\
& \cellcolor{lm_purple_low}\checkmark & \cellcolor{lm_purple_low}\checkmark & \cellcolor{lm_purple_low}\textbf{0.297}{\tiny$\pm$0.45} & \cellcolor{lm_purple_low}0.335 & \cellcolor{lm_purple_low}\textbf{0.536} & \cellcolor{lm_purple_low}0.992{\tiny$\pm$0.01} & \cellcolor{lm_purple_low}0.762{\tiny$\pm$0.25} \\
\midrule
\multirow{3}{*}{Uterus}
& \checkmark & & 0.184{\tiny$\pm$0.32} & \textbf{0.955} & 0.052 & 0.988{\tiny$\pm$0.01} & 0.837{\tiny$\pm$0.21} \\
& & \checkmark & 0.394{\tiny$\pm$0.44} & 0.676 & 0.482 & 0.989{\tiny$\pm$0.01} & 0.836{\tiny$\pm$0.22} \\
& \cellcolor{lm_purple_low}\checkmark & \cellcolor{lm_purple_low}\checkmark & \cellcolor{lm_purple_low}\textbf{0.401}{\tiny$\pm$0.47} & \cellcolor{lm_purple_low}0.424 & \cellcolor{lm_purple_low}\textbf{0.600} & \cellcolor{lm_purple_low}0.989{\tiny$\pm$0.02} & \cellcolor{lm_purple_low}0.806{\tiny$\pm$0.23} \\
\midrule
\multirow{3}{*}{Prostate}
& \checkmark & & 0.069{\tiny$\pm$0.21} & \textbf{0.994} & 0.033 & 0.958{\tiny$\pm$0.05} & 0.900{\tiny$\pm$0.17} \\
& & \checkmark & 0.262{\tiny$\pm$0.42} & 0.923 & 0.299 & 0.959{\tiny$\pm$0.05} & 0.898{\tiny$\pm$0.18} \\
& \cellcolor{lm_purple_low}\checkmark & \cellcolor{lm_purple_low}\checkmark & \cellcolor{lm_purple_low}\textbf{0.465}{\tiny$\pm$0.49} & \cellcolor{lm_purple_low}0.645 & \cellcolor{lm_purple_low}\textbf{0.587} & \cellcolor{lm_purple_low}0.971{\tiny$\pm$0.05} & \cellcolor{lm_purple_low}0.890{\tiny$\pm$0.19} \\
\midrule
\multirow{3}{*}{Breast}
& \checkmark & & 0.152{\tiny$\pm$0.33} & \textbf{0.934} & 0.195 & 0.978{\tiny$\pm$0.03} & 0.772{\tiny$\pm$0.24} \\
& & \checkmark & 0.316{\tiny$\pm$0.45} & 0.602 & 0.544 & 0.980{\tiny$\pm$0.03} & 0.767{\tiny$\pm$0.24} \\
& \cellcolor{lm_purple_low}\checkmark & \cellcolor{lm_purple_low}\checkmark & \cellcolor{lm_purple_low}\textbf{0.395}{\tiny$\pm$0.48} & \cellcolor{lm_purple_low}0.412 & \cellcolor{lm_purple_low}\textbf{0.729} & \cellcolor{lm_purple_low}0.984{\tiny$\pm$0.03} & \cellcolor{lm_purple_low}0.762{\tiny$\pm$0.25} \\
\midrule
\multirow{3}{*}{Cervix}
& \checkmark & & 0.167{\tiny$\pm$0.32} & \textbf{0.932} & 0.145 & 0.993{\tiny$\pm$0.01} & 0.800{\tiny$\pm$0.23} \\
& & \checkmark & 0.337{\tiny$\pm$0.45} & 0.496 & 0.513 & 0.993{\tiny$\pm$0.01} & 0.796{\tiny$\pm$0.23} \\
& \cellcolor{lm_purple_low}\checkmark & \cellcolor{lm_purple_low}\checkmark & \cellcolor{lm_purple_low}\textbf{0.355}{\tiny$\pm$0.47} & \cellcolor{lm_purple_low}0.299 & \cellcolor{lm_purple_low}\textbf{0.632} & \cellcolor{lm_purple_low}0.993{\tiny$\pm$0.01} & \cellcolor{lm_purple_low}0.777{\tiny$\pm$0.24} \\
\bottomrule
\end{tabular}}
\caption{Ablation results on OOD tumor types. Mean $\pm$ std values are reported. Stat Test indicates statistical test, and FP Gating denotes false positive gating.}
\label{tab:ablation_subset}
\end{table}

\subsection{Ablation Study}
To evaluate the effectiveness of components in R$^2$-Seg, ablations were performed on the \textit{statistical testing} and \textit{false-positive gating} modules under identical settings with a frozen \textit{BiomedParse} backbone to ensure fair comparison.
\subsubsection{Effect of Statistical Testing}
As shown in Table~\ref{tab:ablation_subset}, removing the two-sample statistical test substantially increases false activations, particularly in low-contrast organ interiors.  
Without this hypothesis-testing step, 
noisy regions are often misinterpreted as lesions, reducing Dice and precision despite higher sensitivity.
Moreover, reintroducing the MMD-based test with Benjamini–Hochberg correction restores clear boundaries and provides a robust, label-free criterion for reliable confidence estimation.
\subsubsection{Effect of False-Positive Gating}
Disabling the false-positive gating mechanism leads to a significant increase in background activations and a marked drop in specificity across all tumor types.
Although a more permissive activation threshold occasionally improves sensitivity, both Dice and specificity deteriorate sharply.
In contrast, the hierarchical gating strategy in the R$^2$-Seg adaptively suppresses low-confidence regions, maintaining high specificity while preventing over-segmentation.
Overall, the integration of statistical testing and false-positive gating is essential for achieving reliable and well-calibrated test-time adaptation.


\section{Discussion}
\paragraph{Avoiding Catastrophic Forgetting: The Unique Advantage of Training-Free.}

Unlike adaptive methods such as fine-tuning or LoRA that require parameter updates, $R^2$-Seg is entirely training-free. Experimental results clearly demonstrate that while fine-tuning methods may improve Dice scores on out-of-distribution (OOD) tumor data, they exhibit \textit{catastrophic forgetting} on in-distribution normal organ segmentation tasks. $R^2$-Seg fundamentally avoids this problem by keeping the base model (BioMedParse) weights fully frozen. It enhances robustness on OOD tasks without sacrificing the model's original generalization capabilities. 

\paragraph{Reason Before Reject: Statistically Eliminating False Positives.}

The core challenge in OOD tumor segmentation is that base models generate numerous false positives, leading to overdiagnosis. $R^2$-Seg addresses this through a two-stage mechanism: (1) \textit{Reason}: An LLM-guided anatomical planner first restricts the search to anatomically normal and plausible multi-scale ROIs, inherently filtering out substantial background noise. (2) \textit{Reject}: More critically, we introduce a nonparametric two-sample statistical test (MMD Test). This provides a principled rejection mechanism that no longer relies on model-dependent confidence scores but instead tests whether candidate regions are statistically significantly different from normal organs. Ablation experiments (Table~\ref{tab:ablation_subset}) and visualization results (Fig.~\ref{fig:exp_vis}) both confirm that this rejection step is crucial for achieving high specificity and suppressing false positives.

\paragraph{Limitations.} Balancing false positives and false negatives remains a fundamental challenge in tumor segmentation. Under-segmentation, characterized by a high false negative rate, can delay diagnosis and lead to under-treatment, thereby increasing mortality risk. Conversely, over-segmentation, associated with a high false positive rate, may result in inaccurate diagnosis and unnecessary treatment, elevating patient anxiety and financial burden. Like other methods, R$^2$-Seg also exhibits this trade-off. While it effectively suppresses false positives, it risks rejecting true positives and missing early radiological biomarkers. Achieving simultaneous improvement in both sensitivity and specificity remains an open and critical challenge. 

\section{Conclusion}
In this work, we propose R$^2$-Seg, a novel, training-free framework that addresses out-of-distribution (OOD) tumor segmentation by tackling the high false-positive rates of foundation models. Our framework operates on a reason-and-reject principle, first using an LLM-guided planner for anatomical reasoning to define localized regions of interest, and then applying a two-sample statistical test to rigorously reject false-positive candidates generated by a frozen segmentor. Because our method requires no parameter updates, it entirely avoids the catastrophic forgetting associated with fine-tuning. On multi-center and multi-modal OOD benchmarks, R$^2$-Seg substantially improves Dice, specificity, and sensitivity over baselines, offering a robust, deployable, and generalizable test-time pipeline for safer medical tumor parsing.

{
    \small
    \bibliographystyle{ieeenat_fullname}
    \bibliography{main}
}

\clearpage
\setcounter{page}{1}
\setcounter{section}{0}
\maketitlesupplementary

\section{Method Details}
\label{app:method_details}

\subsection{Planner Outputs and ROI Construction}

\subsubsection{LLM Planner and Prompt Normalization}

As illustrated in Algorithm.~\ref{alg:planning}, the LLM planner $\Phi$ transforms a free-form tumor description into structured geometric guidance for ROI construction. Its output consists of anchor organs, ROI parameters, and a concise rationale, each generated through a deterministic template (Appendix~\ref{app:prompts}). The planner interprets the anatomical site implied by the tumor type, identifies nearby organs that are consistently visible, and determines padding, scale jitter, and optional squaring rules for the ROI. All outputs are formatted as JSON to ensure reproducibility and model-agnostic integration.

Because clinical terminology varies across datasets and institutions, user-provided concepts are normalized before being passed to the segmentor. A dedicated LLM normalizer $\Pi$ maps arbitrary descriptions from an open prompt space $\mathcal{C}$ onto a fixed vocabulary $\mathcal{C}^\star$ aligned with BiomedParse training. This process relies on a template that restricts outputs to the predefined vocabulary while allowing the LLM to justify its choice. By enforcing a stable set of canonical terms, the pipeline prevents stylistic drift, maintains consistent text conditioning across all R$^2$-Seg views, and ensures that any variation in predictions arises solely from visual input rather than prompt variability.

Together, the planner and normalizer provide anatomically informed yet text-stable conditioning for downstream segmentation, enabling robust performance across datasets without modifying model parameters.

\subsubsection{Prompt Design}
\label{app:prompts}

The R$^2$-Seg pipeline relies on a small set of deterministic prompts that govern anatomical planning, concept normalization, and segmentation. These prompts standardize how the planner reasons about anatomy and how the segmentor receives textual instructions, thereby isolating visual variability from linguistic variability and improving cross-dataset robustness.

The planner operates on free-form tumor descriptions using a system prompt that instructs the model to consider anatomical context, identify adjacent organs that are reliably detectable, and generate ROI parameters such as padding, scale jitter, and optional squaring rules. The output follows a fixed JSON schema to ensure consistent behavior across tumor types and imaging modalities. A brief user prompt specifies the tumor type and modality, prompting the planner to derive anchors and ROI rules tailored to the anatomical site.

To address the diversity of medical terminology, a dedicated normalizer $\Pi$ maps user-provided concepts to the canonical vocabulary used during BiomedParse training. The normalizer’s template restricts output to a predefined vocabulary list while allowing the LLM to provide justification. This step stabilizes the text encoder and ensures consistent conditioning across all TTA crops.

After anchors and normalized concepts are obtained, segmentation prompts are generated through concise templates. Each anchor organ is passed to the segmentor using the directive \textit{``segment the \textless ORGAN\_NAME\textgreater ''}, a phrasing aligned with the model’s training data. Tumor segmentation within each ROI uses the parallel template \textit{``segment the \textless TUMOR\_SITE\textgreater\ tumor''}. These minimal forms avoid unnecessary modifiers, ensuring that prediction differences across crops reflect only visual context changes rather than variations in phrasing.

Overall, the prompt design emphasizes stability, clarity, and consistency. It separates anatomical reasoning from segmentation, maintains uniform textual conditioning across views, and allows R$^2$-Seg to remain fully training-free while retaining strong cross-domain robustness.

\begin{algorithm}[t]
\caption{Anatomy-aware Planning and ROI Proposal}
\label{alg:planning}
\begin{algorithmic}[1]
\REQUIRE Cancer type $c$, input image $I$, in-plane spacings $(s_x, s_y)$, 
padding $\delta$, and scale jitters $\Gamma$
\vspace{2pt}

\STATE $(\mathcal{A},\, \mathcal{I}_{\mathrm{ROI}},\, r) \leftarrow \Phi(c)$
\STATE \textbf{for each} anchor $a \in \mathcal{A}$:  
\hspace{4mm} $c_a \leftarrow \Pi(a)$ \hfill (normalize anchor name)  
\hspace{4mm} $M_a \leftarrow \mathbb{1}\!\left\{ f_\theta(I;c_a) \ge \tau_a \right\}$ \hfill (anchor mask)

\STATE $M_\cup \leftarrow \bigcup_{a \in \mathcal{A}} M_a$ \hfill (union anchor mask)
\STATE $B_0 \leftarrow \mathsf{BBox}(M_\cup)$

\STATE $(m_x, m_y) \leftarrow 
\left( \left\lceil \delta_x / s_x \right\rceil,\;
       \left\lceil \delta_y / s_y \right\rceil \right)$

\STATE $\widehat{B} \leftarrow \mathsf{Pad}(B_0; m_x, m_y)$
\STATE \textbf{if} squaring is requested: \hspace{3mm}
        $\widehat{B} \leftarrow \mathsf{Square}(\widehat{B})$

\STATE \textbf{return}
        $\{\, B_\gamma = \mathsf{Scale}(\widehat{B}; \gamma)\, \}_{\gamma \in \Gamma}$
        and $M_\cup$

\end{algorithmic}
\end{algorithm}

\subsubsection{Anchor Masks and Base Bounding Box}

For each anchor $a \in \mathcal{A}$, the text-driven segmentor $f_\theta$ generates a probability map $P_a$, which is thresholded at $\tau_a$ to obtain a binary mask $M_a = \mathbb{1}\{P_a \ge \tau_a\}$.  
The union mask $M_\cup = \bigcup_{a\in\mathcal{A}} M_a$ defines an initial region of interest, and its axis-aligned bounding box is denoted by $B_0 = \mathsf{BBox}(M_\cup)$. When anchors are unreliable or missing, a conservative fallback sets $B_0$ to the full image frame so that downstream computations remain well-defined.

\subsubsection{Padding, Squaring, and Multi-scale Jitter}

Let $(s_x, s_y)$ denote the in-plane physical spacings in mm/pixel. The padding vector $\delta = (\delta_x, \delta_y)$ is converted to pixel margins
\[
m_x = \left\lceil \frac{\delta_x}{s_x} \right\rceil,\qquad
m_y = \left\lceil \frac{\delta_y}{s_y} \right\rceil,
\]
and applied to $B_0$ in image coordinates.  
If $\text{square}=\text{true}$, the padded bounding box is expanded along its shorter side to obtain a square ROI. A family of jittered ROIs $\{B_\gamma\}_{\gamma\in\Gamma}$ is then formed by isotropically scaling the side length of this squared box by each $\gamma \in \Gamma$, while keeping its center fixed.

Each ROI $B_\gamma$ induces a crop $I|_{B_\gamma}$ that is segmented independently and later restored to the native image canvas during aggregation.

\subsection{Segmentation and ROI Fusion}

\subsubsection{Transform Set and Inverse Alignment}
Let $\mathcal{G}=\{g_{\text{id}}, g_{\text{lr}}, g_{\text{tb}}\}$ denote the identity, left--right flip, and top--bottom flip transforms.  
For each crop $I|_{B_\gamma}$ and each view $g\in\mathcal{G}$, we compute a prediction
\[
P^{(\gamma,g)} = f_\theta\big(g(I|_{B_\gamma});\, c_{\mathrm{tumor}}\big).
\]
The prediction is then realigned to the native image frame using $\mathsf{Inv}(g)$ and restored to the full canvas based on the known ROI coordinates.  
This ensures that all predictions, regardless of view or crop, are accumulated in a common spatial reference frame.

\subsubsection{Aggregation Across Views and Supports}
Let $A\in\{\max,\text{median},\text{mean}\}$ be an element-wise fusion rule. The fused probability over an ROI is defined as
\[
P^{(\gamma)}(x) 
= A_{g\in\mathcal{G}}
\Big[
  \big(\mathsf{Restore}_{B_\gamma}\circ \mathsf{Inv}(g)\big)\,P^{(\gamma,g)}
\Big](x).
\]
When restored ROIs overlap, contributions are averaged to avoid bias from differences in ROI coverage.  
In addition to ROI-level predictions, we compute a full-frame prediction $P^{\mathrm{full}}$ (with the same test-time augmentations).  
To preserve confident detections that may appear only at the global scale or only within a jittered ROI, we adopt a conservative support fusion:
\begin{equation}
\label{eq:roi_full_fusion}
P^{\mathrm{final}}(x) 
= \max\Big( P^{\mathrm{full}}(x),\ \max_{\gamma\in\Gamma} P^{(\gamma)}(x) \Big).
\end{equation}
This rule retains any high-probability region observed in either support while still exploiting the anatomical focus provided by the ROI crops.  
Empirically, using $A=\max$ or $A=\text{median}$ yields stable results, whereas $\text{mean}$ may be preferred when suppressing isolated peaks is desirable.

\begin{algorithm}[t]
\caption{Segmentation and ROI Fusion}
\label{alg:seg}
\begin{algorithmic}[1]
\REQUIRE Image $I$, ROIs $\{B_\gamma\}$, transforms $\mathcal{G}$, fusion rule $A$
\STATE $P^{\mathrm{full}}\leftarrow A_{g\in\mathcal{G}}\big[\mathsf{Inv}(g)\,f_\theta(g(I);c_{\mathrm{tumor}})\big]$
\FOR{$\gamma\in\Gamma$}
  \FOR{$g\in\mathcal{G}$}
    \STATE $Q\leftarrow f_\theta\big(g(I|_{B_\gamma});\,c_{\mathrm{tumor}}\big)$
    \STATE $Q\leftarrow \mathsf{Inv}(g)(Q)$; \quad $P^{(\gamma,g)}\leftarrow \mathsf{Restore}_{B_\gamma}(Q)$
  \ENDFOR
  \STATE $P^{(\gamma)}\leftarrow A_{g\in\mathcal{G}}[P^{(\gamma,g)}]$ with overlap-count normalization
\ENDFOR
\STATE \textbf{return} $P^{\mathrm{final}}=\max\big(P^{\mathrm{full}},\ \max_{\gamma} P^{(\gamma)}\big)$
\end{algorithmic}
\end{algorithm}

\subsubsection{Monotonicity Property}
Let $\tau>0$ be a binarization threshold, and define the super-level set $\mathcal{S}(P,\tau)=\{x : P(x)\ge\tau\}$.  
By construction of the fusion rule,
\[
\begin{aligned}
\mathcal{S}\big(P^{\mathrm{final}},\tau\big)
&= \mathcal{S}\big(\max(P^{\mathrm{full}},\max_\gamma P^{(\gamma)}),\tau\big)
\\
&\supseteq \mathcal{S}\big(P^{\mathrm{full}},\tau\big)\ \cup\ \bigcup_{\gamma}\mathcal{S}\big(P^{(\gamma)},\tau\big).
\end{aligned}
\]

Thus no connected component captured by any support is removed when fusing predictions at the same threshold.  
This property favors recall, while precision is deferred to later statistical screening steps.
\subsection{Candidate Extraction and Cross-slice Linking}

From the fused probability map $P^{\mathrm{final}}$, a binary mask
\[
\mathcal{M}=\mathbb{1}\{P^{\mathrm{final}}\ge\tau_{\mathrm{bin}}\}
\]
is obtained and decomposed into 2D connected components $\{C_k\}_{k\in\mathcal{K}}$ using 8-connectivity.  
Small components are removed, and basic region descriptors (centroid, area, mean probability) are retained for later statistical screening.  

\subsection{Statistical Screening with MMD and FDR Control}

\subsubsection{Feature sampling and kernel choice}

For each candidate $C_k$ and the control region $M_\cup$, pixel-level features $\phi$ are sampled uniformly up to a cap of $4\mathrm{k}$ points per set.  
We use a Gaussian kernel with bandwidth $\sigma$ determined by the median heuristic on the pooled sample.  
These implementation details complement the high-level procedure given in the main paper.

\subsubsection{Unbiased MMD and permutation testing}
We use the unbiased MMD estimator and permutation test exactly as defined in the main text (Sec.~\ref{sec:stats}); here we detail the sampling caps, bandwidth selection, and smoothed counting for $p$-values.
Let $X$ and $Y$ denote the sampled features from $C_k$ and $M_\cup$.  
The unbiased MMD estimator is used, and permutation testing is performed by recomputing the statistic on $B$ random shuffles of the pooled sample.  
The $p$-value estimate
\[
p_k = \frac{\#\{b:\widehat{\mathrm{MMD}}^2_{\mathrm{perm},b}\ge \widehat{\mathrm{MMD}}^2_{\mathrm{obs}}\}+1}{B+1}
\]
includes the standard smoothing factor to avoid zero-valued estimates.  
All computations are vectorized and can be executed on GPU for efficiency.

\subsubsection{Multiple testing}

Across all candidates, the Benjamini–Hochberg procedure at level $\alpha$ determines the subset declared significant.  
Let $p_{(1)}\le\cdots\le p_{(|\mathcal K|)}$ be the ordered $p$-values; the decision index
\[
i^\ast=\max\{i:\, p_{(i)}\le \alpha\,i/|\mathcal K|\}
\]
selects the retained components $\{C_{(1)},\dots,C_{(i^\ast)}\}$.  
The pipeline remains compatible with alternative corrections when sample sizes are small.

\begin{algorithm}[t]
\caption{Candidate-level screening with MMD and BH-FDR}
\label{alg:stats}
\begin{algorithmic}[1]
\REQUIRE Candidates $\{C_k\}$, control mask $M_\cup$, features $\phi$, permutations $B$, FDR level $\alpha$
\FOR{each $k$}
  \STATE Sample $X\subseteq\phi(I|_{C_k})$ and $Y\subseteq\phi(I|_{M_\cup})$ with caps
  \STATE Set kernel bandwidth $\sigma$ via median heuristic
  \STATE Compute $\widehat{\mathrm{MMD}}^2_{\mathrm{obs}}(X,Y)$
  \STATE Estimate $p_k$ by $B$ permutations with smoothed counts
\ENDFOR
\STATE Apply BH-FDR at level $\alpha$ and retain significant candidates
\STATE \textbf{return} filtered candidate set
\end{algorithmic}
\end{algorithm}

\subsection{False-positive Gating for Empty-mask Cases}
\label{app:empty_mask_gating}

Let $P^{\mathrm{final}}$ denote the fused probability map defined in Eq.~\eqref{eq:roi_full_fusion}, where predictions within each view set $\mathcal{G}$ are aggregated by mean, and ROI-vs-full fusion uses the $\max$ rule.
We apply a three-level gating strategy complementary to statistical testing:

\paragraph{(L1) Existence gate.}
Compute the global maximum $p_{\max}=\max_x P^{\mathrm{final}}(x)$, the positive-pixel ratio within the ROI domain
\[
\rho=\frac{|\{x\in\Omega_{\mathrm{ROI}}:P^{\mathrm{final}}(x)\ge\tau_{\mathrm{bin}}\}|}{|\Omega_{\mathrm{ROI}}|},
\]
and a KS statistic $p_{\mathrm{KS}}$ comparing foreground and background probabilities inside $\Omega_{\mathrm{ROI}}$.
If $p_{\max}<\tau_{\max}$ or $\rho<\tau_{\rho}$ or $p_{\mathrm{KS}}>\tau_{\text{KS}}$, the case is declared negative and no candidates are extracted.

\paragraph{(L2) Candidate-level gate.}
For candidates $\{C_k\}$ obtained from $P^{\mathrm{final}}$, require
\[
|C_k|\ge A_{\min},\qquad
\overline{P}_k\ge \tau_{\text{mean}},\qquad
\frac{|C_k\cap M_{\cup}|}{|C_k|}\ge \tau_{\cap},
\]
where $\overline{P}_k$ is the mean probability over $C_k$ and $M_{\cup}$ is the organ control region.

\paragraph{(L3) Case-level score.}
Define $S_k=\overline{P}_k\sqrt{|C_k|}$ and $S^\ast=\max_k S_k$.
If $S^\ast<\tau_{\text{case}}$, return an all-zero mask.
This gate acts as a conservative post-hoc calibration when empty cases dominate the test set.

\subsection{Computational Aspects and Approximations}

Let $|\mathcal{K}|$ denote the number of candidates, and let $m$ and $n$ be the sampled pixels from each candidate and the control region (both capped).  
With $B$ permutations, the screening cost scales as $O(|\mathcal{K}|\,B\,(m+n)^2)$, with vectorized distance computations giving a modest constant factor.  
If necessary, approximations such as random Fourier features or Nyström sampling can reduce the quadratic kernel cost while preserving decision quality and FDR validity.  
All steps are training-free and do not update model parameters at test time.

\subsection{Aggregation Rules and Robustness}

The ROI fusion rule in \eqref{eq:roi_full_fusion} ensures that high-confidence signals from either support are preserved, improving recall on challenging cases.  
The choice of view-aggregation rule $A$ modulates the sharpness of the fused map: $\max$ preserves peaks, $\text{median}$ stabilizes against outlier views, and $\text{mean}$ suppresses noise but may attenuate small structures.  
Combined with statistical screening and conservative gating, this design yields a robust balance between sensitivity and specificity under distribution shift.

\section{Implementation Details}
\label{supply:implementation}
\subsection{Parameters}
We categorize all hyper-parameters into three coherent categories—\textit{scoring}, \textit{statistical}, and \textit{geometric/gating}. 
A small unlabeled development split, strictly disjoint from the test data, is used only once to set a few global defaults. 
All remaining thresholds are derived directly from instance-level prediction statistics at test time; thus, \textit{no supervision} is involved to select parameters.

\begin{figure*}
    \centering
    \includegraphics[width=1\linewidth]{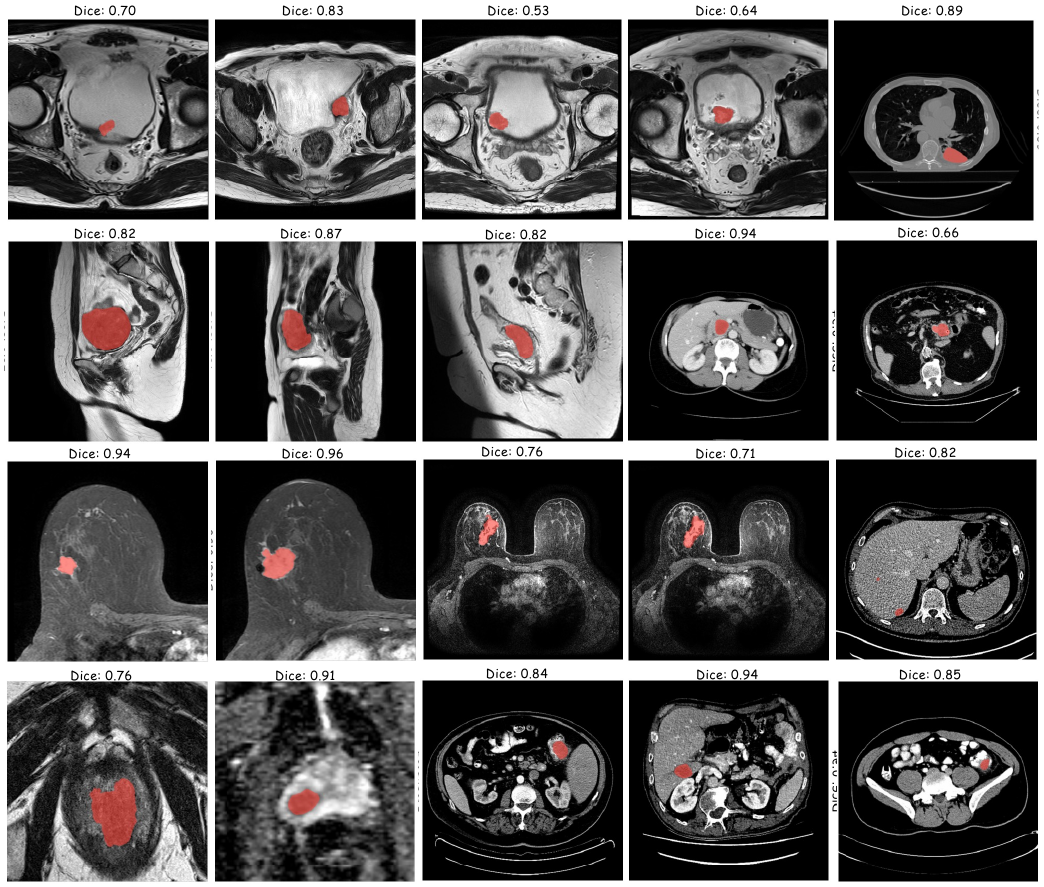}
    \caption{More qualitative results predicted by our R$^2$-Seg. Our model is capable of recognizing tumors of varying sizes, across different scanning planes and organs, under diverse fields of view.}
    \label{fig:app_vis}
\end{figure*}

\subsubsection{Scoring thresholds.} 
The test-time augmentation scheme uses max fusion to retain peak responses while suppressing view-specific noise. 
Scale jitter is kept moderate (\([0.8,1.0,1.2]\)). 
The binarization threshold $\tau_{\mathrm{bin}}$ is chosen once on the unlabeled development split and thereafter fixed. 
All other scoring-related quantities operate directly on each image’s prediction statistics, ensuring a completely label-free procedure at test time.
Empirically, $\tau_{\mathrm{bin}}{=}0.4$ balances recall and precision consistently across organs and imaging planes.

\subsubsection{Statistical thresholds.}
All statistical decisions rely solely on prediction-derived feature distributions computed inside the ROI. 
Two-sample testing and FDR control aim to bound false discoveries based on these distributions. 
Here, the \textit{Type-I error} denotes falsely rejecting the null hypothesis—i.e., misidentifying a normal region as abnormal—when their predicted feature distributions are statistically similar. 
To control this error in a label-free manner, we adopt the median heuristic for the Gaussian kernel bandwidth, cap each region at $4\mathrm{k}$ samples, and set a single Benjamini–Hochberg level $\alpha$ using only the development split. 
All statistical thresholds remain fixed at test time and operate exclusively on model-produced statistics.

\subsubsection{Geometric and gating thresholds.} 
Geometric and gating decisions also rely entirely on statistics computed from the model’s predictions within each ROI. 
The existence gate evaluates the positive-pixel ratio $r$ and ROI-max probability $m$, where reference levels $\tau_{\mathrm{ratio}}$ and $\tau_{\max}$ are fixed percentiles estimated from the development split and do not require any supervision. 
The morphology gate constrains the candidate area and confidence via low-percentile lesion priors ($\tau_{\mathrm{area}}$, $\tau_{\mathrm{mean}}$), computed from model output statistics. 
The control-overlap gate enforces anatomical consistency using a modest $\tau_{\mathrm{inter}}{=}0.05$, or an adaptive form $\tau_{\mathrm{inter}}\!\leftarrow\!c\rho$ with $c{\approx}0.05$. 
At the case level, the stability score $S_k\!=\!\overline{P}_k\sqrt{|C_k|}$ is compared with a global cutoff $\tau_{\mathrm{case}}{=}2.0$, chosen to suppress false positives without requiring any supervision. 
ROI padding is defined in millimetres for modality independence; $25$\,mm corresponds to the 95th percentile of anchor-to-lesion distances in abdominal and pelvic scans.

\subsubsection{Default settings and sensitivity.} 
Unless otherwise specified, we use fixed defaults derived from the development split: aggregation=max, $\tau_{\mathrm{bin}}{=}0.4$, $\alpha{=}0.05$, $\tau_{\max}{=}0.45$, $\tau_{\mathrm{pos}}{=}0.45$, $\tau_{\mathrm{ratio}}{=}2{\times}10^{-4}$, $\tau_{\mathrm{area}}{=}80$\,px, $\tau_{\mathrm{mean}}{=}0.5$, $\tau_{\mathrm{inter}}{=}0.05$, $\tau_{\mathrm{case}}{=}2.0$, ROI padding $25$\,mm, and at most $4\mathrm{k}$ samples per region. 
Ablation studies show that moderate (±20\%) variations around these defaults marginally affect Dice, with $\tau_{\mathrm{bin}}$ and $\tau_{\mathrm{case}}$ having the largest influence. 
In practice, adjusting only these two parameters achieves near-optimal performance while keeping all statistical thresholds and ROI padding fixed.

\subsubsection{Computational burden}
When finetuning BiomedParse with LoRA, the bakcbone was kept frozen while the pixel-decoder was updated for 5 epochs following its official settings. This process required approximately 20 hours on an NVIDIA A100 GPU (80 GB memory). In contrast, R$^2$-Seg requires no additional training or finetuning. Its inference was performed on single Nvidia RTX4090 GPU with an average inference time of approximately 4 seconds per image.

\begin{table*}[!ht]
\centering
\small
\setlength{\tabcolsep}{4pt}
\resizebox{0.75\linewidth}{!}{
\begin{tabular}{l l c c c c c c}
\toprule
Tumor & Method & Dice & Sensitivity & Specificity & Accuracy & CA \\
\midrule
\multirow{4}{*}{Bladder tumor} & BiomedParse & 0.069 & 1.000 & 0.000 & 0.976 & 0.546 \\
& BiomedParse-FT & 0.482 & 0.992 & 0.254 & 0.995 & 0.687 \\
& BiomedParse-LoRA & 0.578 & 0.960 & 0.456 & 0.996 & 0.677 \\
& R$^2$-Seg (Ours) & 0.297 & 0.335 & 0.536 & 0.992 & 0.762 \\
\midrule
\multirow{4}{*}{Breast tumor} & BiomedParse & 0.144 & 0.998 & 0.030 & 0.952 & 0.582 \\
& BiomedParse-FT & 0.559 & 0.982 & 0.411 & 0.995 & 0.688 \\
& BiomedParse-LoRA & 0.611 & 0.954 & 0.520 & 0.996 & 0.685 \\
& R$^2$-Seg (Ours) & 0.395 & 0.412 & 0.728 & 0.984 & 0.762 \\
\midrule
\multirow{4}{*}{Cervix tumor} & BiomedParse & 0.154 & 1.000 & 0.000 & 0.985 & 0.598 \\
& BiomedParse-FT & 0.424 & 0.991 & 0.197 & 0.995 & 0.699 \\
& BiomedParse-LoRA & 0.485 & 0.949 & 0.359 & 0.996 & 0.686 \\
& R$^2$-Seg (Ours) & 0.355 & 0.299 & 0.632 & 0.993 & 0.777 \\
\midrule
\multirow{4}{*}{Colon tumor} & BiomedParse & 0.351 & 1.000 & 0.042 & 0.994 & 0.674 \\
& BiomedParse-FT & 0.497 & 0.987 & 0.271 & 0.998 & 0.683 \\
& BiomedParse-LoRA & 0.498 & 0.970 & 0.309 & 0.998 & 0.675 \\
& R$^2$-Seg (Ours) & 0.479 & 0.110 & 0.919 & 0.997 & 0.758 \\
\midrule
\multirow{4}{*}{Kidney tumor} & BiomedParse & 0.138 & 1.000 & 0.067 & 0.990 & 0.553 \\
& BiomedParse-FT & 0.567 & 0.967 & 0.475 & 0.998 & 0.617 \\
& BiomedParse-LoRA & 0.537 & 0.961 & 0.434 & 0.998 & 0.617 \\
& R$^2$-Seg (Ours) & 0.567 & 0.371 & 0.778 & 0.997 & 0.873 \\
\midrule
\multirow{4}{*}{Liver tumor} & BiomedParse & 0.359 & 0.998 & 0.136 & 0.994 & 0.647 \\
& BiomedParse-FT & 0.529 & 0.932 & 0.413 & 0.998 & 0.640 \\
& BiomedParse-LoRA & 0.556 & 0.950 & 0.451 & 0.998 & 0.643 \\
& R$^2$-Seg (Ours) & 0.625 & 0.342 & 0.866 & 0.996 & 0.853 \\
\midrule
\multirow{4}{*}{Lung tumor} & BiomedParse & 0.158 & 0.977 & 0.072 & 0.995 & 0.562 \\
& BiomedParse-FT & 0.393 & 0.982 & 0.195 & 0.998 & 0.652 \\
& BiomedParse-LoRA & 0.460 & 0.915 & 0.376 & 0.998 & 0.640 \\
& R$^2$-Seg (Ours) & 0.446 & 0.071 & 0.884 & 0.997 & 0.752 \\
\midrule
\multirow{4}{*}{Pancreas tumor} & BiomedParse & 0.212 & 1.000 & 0.078 & 0.998 & 0.584 \\
& BiomedParse-FT & 0.422 & 0.955 & 0.354 & 0.999 & 0.576 \\
& BiomedParse-LoRA & 0.417 & 0.980 & 0.328 & 0.999 & 0.584 \\
& R$^2$-Seg (Ours) & 0.739 & 0.131 & 0.935 & 0.999 & 0.896 \\
\midrule
\multirow{4}{*}{Prostate tumor} & BiomedParse & 0.047 & 1.000 & 0.000 & 0.910 & 0.552 \\
& BiomedParse-FT & 0.377 & 0.917 & 0.352 & 0.991 & 0.561 \\
& BiomedParse-LoRA & 0.428 & 0.852 & 0.434 & 0.992 & 0.555 \\
& R$^2$-Seg (Ours) & 0.465 & 0.645 & 0.587 & 0.971 & 0.890 \\
\midrule
\multirow{4}{*}{Uterus tumor} & BiomedParse & 0.190 & 1.000 & 0.000 & 0.977 & 0.632 \\
& BiomedParse-FT & 0.478 & 0.961 & 0.273 & 0.995 & 0.682 \\
& BiomedParse-LoRA & 0.514 & 0.897 & 0.394 & 0.995 & 0.670 \\
& R$^2$-Seg (Ours) & 0.401 & 0.424 & 0.600 & 0.989 & 0.806 \\
\bottomrule
\end{tabular}}
\caption{Mean metrics (all classes included) grouped by tumor type and method.}
\label{tab:tumor_metrics_all_classes}
\end{table*}

\section{Additional Results}

\subsection{Quantitative Results}
The complete quantitative results on all the categories are shown in Table.~\ref{tab:tumor_metrics_all_classes} and Table.~\ref{tab:ablation_metrics}.

\begin{table*}[t]
\centering
\small
\setlength{\tabcolsep}{5pt}

\resizebox{0.8\linewidth}{!}{
\begin{tabular}{l l c c c c c}
\toprule
Tumor & Method & Dice & Sensitivity & Specificity & Accuracy & CA \\
\midrule
\multirow{4}{*}{Bladder tumor} 
& R$^2$-Seg         & 0.297\tiny{$\pm$ 0.449} & 0.335 & 0.536 & 0.992\tiny{$\pm$ 0.013} & 0.762\tiny{$\pm$ 0.245} \\
& R$^2$-Seg wo ST    & 0.257\tiny{$\pm$ 0.419} & 0.540 & 0.419 & 0.990\tiny{$\pm$ 0.015} & 0.770\tiny{$\pm$ 0.241} \\
& R$^2$-Seg wo FPG     & 0.099\tiny{$\pm$ 0.265} & 0.923 & 0.089 & 0.989\tiny{$\pm$ 0.015} & 0.774\tiny{$\pm$ 0.239} \\
\midrule
\multirow{4}{*}{Breast tumor} 
& R$^2$-Seg         & 0.395\tiny{$\pm$ 0.477} & 0.412 & 0.729 & 0.984\tiny{$\pm$ 0.028} & 0.762\tiny{$\pm$ 0.246} \\
& R$^2$-Seg wo ST    & 0.316\tiny{$\pm$ 0.446} & 0.602 & 0.544 & 0.980\tiny{$\pm$ 0.030} & 0.767\tiny{$\pm$ 0.243} \\
& R$^2$-Seg wo FPG     & 0.152\tiny{$\pm$ 0.325} & 0.934 & 0.195 & 0.978\tiny{$\pm$ 0.030} & 0.772\tiny{$\pm$ 0.240} \\
\midrule
\multirow{4}{*}{Cervix tumor} 
& R$^2$-Seg         & 0.355\tiny{$\pm$ 0.465} & 0.299 & 0.632 & 0.993\tiny{$\pm$ 0.008} & 0.777\tiny{$\pm$ 0.240} \\
& R$^2$-Seg wo ST    & 0.337\tiny{$\pm$ 0.446} & 0.496 & 0.513 & 0.993\tiny{$\pm$ 0.007} & 0.796\tiny{$\pm$ 0.230} \\
& R$^2$-Seg wo FPG     & 0.167\tiny{$\pm$ 0.322} & 0.932 & 0.145 & 0.993\tiny{$\pm$ 0.007} & 0.800\tiny{$\pm$ 0.225} \\
\midrule
\multirow{4}{*}{Colon tumor} 
& R$^2$-Seg         & 0.479\tiny{$\pm$ 0.494} & 0.110 & 0.919 & 0.997\tiny{$\pm$ 0.006} & 0.758\tiny{$\pm$ 0.247} \\
& R$^2$-Seg wo ST    & 0.460\tiny{$\pm$ 0.490} & 0.174 & 0.852 & 0.997\tiny{$\pm$ 0.007} & 0.764\tiny{$\pm$ 0.245} \\
& R$^2$-Seg wo FPG     & 0.238\tiny{$\pm$ 0.385} & 0.767 & 0.309 & 0.996\tiny{$\pm$ 0.008} & 0.782\tiny{$\pm$ 0.231} \\
\midrule
\multirow{4}{*}{Kidney tumor} 
& R$^2$-Seg         & 0.567\tiny{$\pm$ 0.490} & 0.371 & 0.778 & 0.997\tiny{$\pm$ 0.008} & 0.873\tiny{$\pm$ 0.215} \\
& R$^2$-Seg wo ST   & 0.477\tiny{$\pm$ 0.490} & 0.723 & 0.628 & 0.994\tiny{$\pm$ 0.011} & 0.887\tiny{$\pm$ 0.204} \\
& R$^2$-Seg wo FPG     & 0.229\tiny{$\pm$ 0.400} & 0.990 & 0.248 & 0.994\tiny{$\pm$ 0.010} & 0.904\tiny{$\pm$ 0.186} \\
\midrule
\multirow{4}{*}{Liver tumor} 
& R$^2$-Seg         & 0.625\tiny{$\pm$ 0.465} & 0.342 & 0.866 & 0.996\tiny{$\pm$ 0.013} & 0.853\tiny{$\pm$ 0.219} \\
& R$^2$-Seg wo ST    & 0.639\tiny{$\pm$ 0.453} & 0.560 & 0.764 & 0.996\tiny{$\pm$ 0.015} & 0.892\tiny{$\pm$ 0.188} \\
& R$^2$-Seg wo FPG     & 0.412\tiny{$\pm$ 0.443} & 0.941 & 0.291 & 0.996\tiny{$\pm$ 0.015} & 0.928\tiny{$\pm$ 0.140} \\
\midrule
\multirow{4}{*}{Lung tumor} 
& R$^2$-Seg         & 0.446\tiny{$\pm$ 0.496} & 0.071 & 0.884 & 0.997\tiny{$\pm$ 0.005} & 0.752\tiny{$\pm$ 0.249} \\
& R$^2$-Seg wo ST   & 0.381\tiny{$\pm$ 0.484} & 0.101 & 0.751 & 0.996\tiny{$\pm$ 0.006} & 0.753\tiny{$\pm$ 0.249} \\
& R$^2$-Seg wo FPG     & 0.209\tiny{$\pm$ 0.395} & 0.614 & 0.379 & 0.996\tiny{$\pm$ 0.006} & 0.758\tiny{$\pm$ 0.246} \\
\midrule
\multirow{4}{*}{Pancreas tumor} 
& R$^2$-Seg         & 0.739\tiny{$\pm$ 0.435} & 0.131 & 0.935 & 0.999\tiny{$\pm$ 0.003} & 0.896\tiny{$\pm$ 0.201} \\
& R$^2$-Seg wo ST    & 0.697\tiny{$\pm$ 0.454} & 0.237 & 0.865 & 0.999\tiny{$\pm$ 0.003} & 0.901\tiny{$\pm$ 0.196} \\
& R$^2$-Seg wo FPG     & 0.288\tiny{$\pm$ 0.439} & 0.803 & 0.304 & 0.999\tiny{$\pm$ 0.003} & 0.911\tiny{$\pm$ 0.184} \\
\midrule
\multirow{4}{*}{Prostate tumor} 
& R$^2$-Seg         & 0.465\tiny{$\pm$ 0.486} & 0.645 & 0.587 & 0.971\tiny{$\pm$ 0.050} & 0.890\tiny{$\pm$ 0.190} \\
& R$^2$-Seg wo ST    & 0.262\tiny{$\pm$ 0.418} & 0.923 & 0.299 & 0.959\tiny{$\pm$ 0.052} & 0.898\tiny{$\pm$ 0.175} \\
& R$^2$-Seg wo FPG     & 0.069\tiny{$\pm$ 0.208} & 0.994 & 0.033 & 0.958\tiny{$\pm$ 0.051} & 0.900\tiny{$\pm$ 0.171} \\
\midrule
\multirow{4}{*}{Uterus tumor} 
& R$^2$-Seg         & 0.401\tiny{$\pm$ 0.466} & 0.424 & 0.600 & 0.989\tiny{$\pm$ 0.016} & 0.806\tiny{$\pm$ 0.234} \\
& R$^2$-Seg wo ST    & 0.394\tiny{$\pm$ 0.444} & 0.676 & 0.482 & 0.989\tiny{$\pm$ 0.014} & 0.836\tiny{$\pm$ 0.215} \\
& R$^2$-Seg wo FPG     & 0.184\tiny{$\pm$ 0.324} & 0.955 & 0.052 & 0.988\tiny{$\pm$ 0.014} & 0.837\tiny{$\pm$ 0.213} \\
\bottomrule
\end{tabular}}
\caption{Ablation study: mean $\pm$ std across metrics per tumor and method. ST and FPG refer to statistical tests and false positive gating, respectively.}
\label{tab:ablation_metrics}
\end{table*}

\subsection{Visual Results}
\begin{figure*}
    \centering
    \includegraphics[width=1\linewidth]{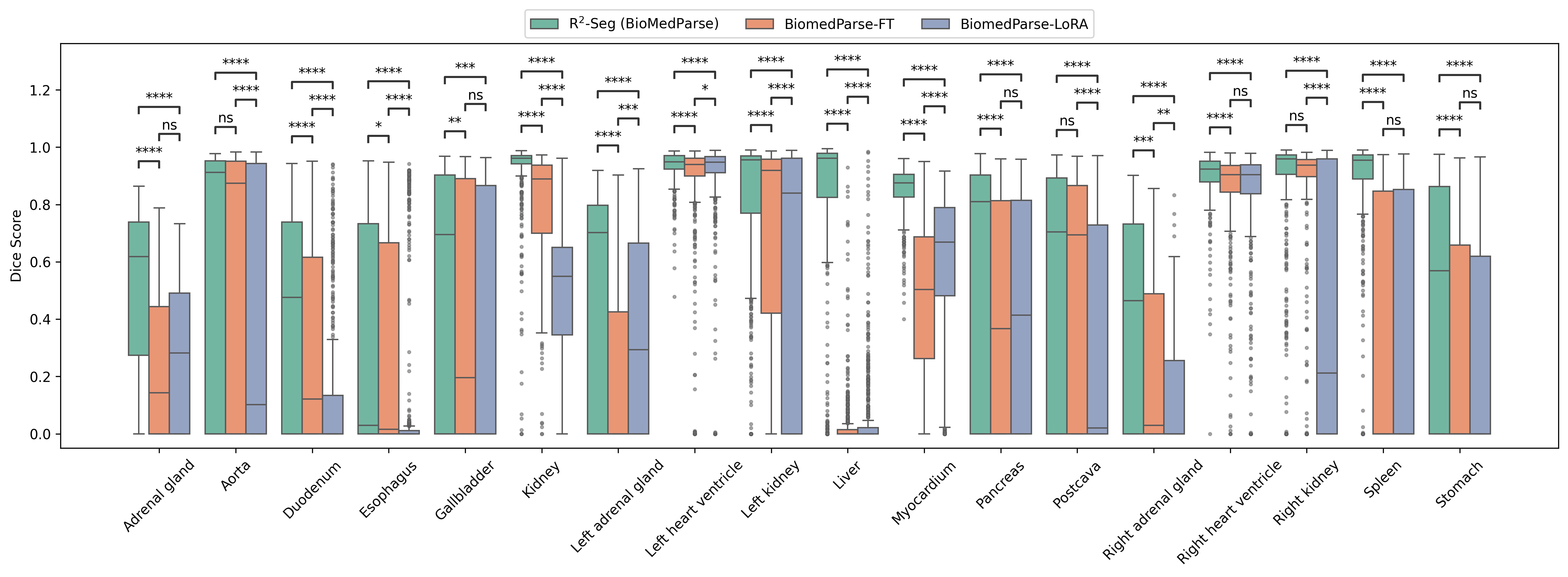}
    \caption{Evaluation of knowledge forgetting on in-distribution MR slices. Statistical tests show that, even finetuned by LoRA, BiomedParse still presents significant performance drops across all organs.}
    \label{fig:MR_abdomen_seg}
\end{figure*}

More experiment results are shown in Fig.~\ref{fig:app_vis} and Fig.~\ref{fig:MR_abdomen_seg}.

\end{document}